\documentclass[11pt,a4paper]{article}
\usepackage{amsmath}
\usepackage{amsfonts}
\usepackage{amssymb}

\usepackage{graphicx}
\usepackage{algorithm}
\usepackage{algorithmic}
\usepackage{booktabs}
\usepackage{microtype}
\usepackage{authblk}
\usepackage{subcaption}
\usepackage[utf8]{inputenc}
\usepackage{bm}
\usepackage[margin=2.5cm]{geometry}
\usepackage{tikz}
\definecolor{darkgreen}{RGB}{0,70,0}
\usepackage[colorlinks=true, citecolor=darkgreen, linkcolor=darkgreen, urlcolor=darkgreen]{hyperref}
\usepackage[numbers]{natbib}
\usepackage[margin=2.5cm]{geometry}
\usetikzlibrary{shapes.geometric, arrows, calc, shapes.multipart, shapes.symbols}
\usetikzlibrary{positioning}
\usetikzlibrary{fit}
\usetikzlibrary{backgrounds}

\begin{document}

 \begin{center}
 \rule{\linewidth}{0.5pt} \\[0.1cm]
 {\LARGE \bfseries Localized Weather Prediction Using Kolmogorov-Arnold Network-Based Models and Deep RNNs} \\[0.1cm]
 \rule{\linewidth}{0.5pt}
 \end{center}




 \begin{center}
\textbf{Ange-Clément Akazan$^{1,2}$},
\textbf{Verlon Roel Mbingui$^{1,2}$},
\textbf{Gnankan Landry Regis N'guessan$^{1,2}$} \textbf{ and }
\textbf{Issa Karambal } \\[0.2cm]
$^1$\text{$\Sigma \eta $igm$\alpha$} Research Group \\
$^2$African Institute for Mathematical Sciences (AIMS), Research and Innovation Centre (RIC), Kigali, Rwanda
\end{center}

\vspace{1cm}


\date{\today}



\begin{abstract}

Weather forecasting is crucial for managing risks and economic planning, particularly in tropical Africa, where extreme events severely impact livelihoods. Yet, existing forecasting methods often struggle with the region's complex, non-linear weather patterns. This study benchmarks deep recurrent neural networks such as LSTM, GRU, BiLSTM, BiGRU, and Kolmogorov-Arnold-based models (KAN and TKAN) for daily forecasting of temperature, precipitation, and pressure in two tropical cities: Abidjan, Côte d'Ivoire (Ivory Coast) and Kigali (Rwanda). We further introduce two customized variants of TKAN that replace its original SiLU activation function with GeLU and MiSH, respectively. Using station-level meteorological data spanning from 2010 to 2024, we evaluate all the models on standard regression metrics. KAN achieves temperature prediction ($R^2=0.9986$ in Abidjan, $0.9998$ in Kigali, MSE $< 0.0014~^\circ$C$^2$), while TKAN variants minimize absolute errors for precipitation forecasting in low-rainfall regimes. The customized TKAN models demonstrate improvements over the standard TKAN across both datasets. Classical RNNs remain highly competitive for atmospheric pressure ($R^2 \approx 0.83{-}0.86$), outperforming KAN-based models in this task. These results highlight the potential of spline-based neural architectures for efficient and data-efficient forecasting.

\paragraph{Keywords: } Weather prediction, Recurrent Neural Networks, Kolmogorov-Arnold Network, Time Series
\end{abstract}

\section{Introduction}
Weather comprises daily atmospheric phenomena such as temperature, precipitation, wind, pressure, humidity, and cloud formation. These variables vary considerably across space and time, making weather forecasting essential, particularly in regions, where climate extreme events pose significant risks to human life and economic development \citep{graves1,kamer_2022}. 
The agricultural sector in developing countries, heavily dependent on weather conditions, is especially vulnerable because many farmers rely on intuition rather than scientific tools, leaving communities unprepared for erratic rainfall, droughts, or floods. In this context, accurate, localized weather forecasting is an urgent enabler of climate resilience, early warning systems, and agricultural planning.
\newline  \newline
Modern weather forecasting combines ground and satellite data with Numerical Weather Prediction (NWP) models to simulate and predict atmospheric patterns accurately \citep{79859,https://doi.org/10.1029/RG024i004p00701,waqas2024artificial}.
 However, modeling the atmosphere is inherently difficult due to its chaotic and complex nature, and traditional physics-based models often lack the flexibility to capture atmospheric randomness, leading to modeling errors \cite{waqas2024artificial,bochenek2022machine}. Even with perfect initial data, Numerical Weather Prediction (NWP) models can omit key atmospheric characteristics, and their reliance on computationally intensive numerical solutions requires the use of supercomputers \citep{ren2021deep}.
To address these limitations, statistical time series methods have been employed as an alternative offering greater flexibility, faster computation, and the ability to capture temporal dependencies in weather data. Common approaches include the Autoregressive Moving Average (ARMA) model \citep{guiot1986arma}, the Autoregressive Integrated Moving Average (ARIMA) model \citep{sawaragi1979statistical, greenland1989climate}, and the Autoregressive Conditional Heteroskedasticity (ARCH) model \citep{tol1997autoregressive}. Although these methods provide useful information, they often struggle to capture complex non-linear temporal dynamics, as is often the case in climate and weather time series data, as noted by \cite{ZHANG2023143}. 
\newline \newline
In recent years, deep Recurrent Neural Networks (RNNs) such as LSTM \cite{artr}, GRU \cite{cho-etal-2014-properties}, BiLSTM, and BiGRU \cite{bin2018describing,Su_2019,cheng2019data} have gained popularity for weather prediction due to their ability to capture complex temporal dependencies in sequential data \cite{hassan2023machine,chhetri2020deep}. However, they are slow to train due to limited parallelization, prone to gradient issues, especially when facing long-term dependencies in data.
Unlike deep learning approach based on fixed activation functions, Kolmogorov-Arnold Networks (KANs), as introduced in  \citep{liu2024kan}, approximate hidden representation  by training univariate B-splines, thus enabling smoother, more localized, and expressivity with fewer parameters \citep{Gao2025, Alves2024}. Given these properties, KAN has the potential to learn useful long-term dependencies in data.
Its  temporal extension, TKAN \citep{genet2024tkan, Sarkar2025}, incorporates memory mechanisms, offering robust, interpretable, and data-efficient solutions for complex weather forecasting tasks involving nonlinearity, sparsity, and long-term temporal dependencies \citep{LiuKANI2025}. Together, these features position KANs and TKANs as powerful and interpretable tools for advancing localized, and data-efficient weather forecasting, and valuable alternatives to deep RNNs.
\newline \newline
This study aims to evaluate the potential of Kolmogorov-Arnold Networks (KANs) and their temporal variant (Temporal KANs) for localized short-term weather prediction (1-day prediction) in two African capital cities: Abidjan (Côte d'Ivoire) and Kigali (Rwanda).  We focus on forecasting three key atmospheric variables, precipitation, temperature at 2 meters, and surface pressure, to evaluate the models' ability to capture complex and variate temporal patterns in weather data. The performance of KAN-based models will be systematically compared against established deep recurrent neural network benchmarks using a suite of evaluation metrics, including  Mean Squared Error (MSE), Root Mean Squared Error (RMSE), Mean Absolute Error (MAE), the coefficient of determination ($R^2$), and Mean Absolute Percentage Error (MAPE) \citep{paparrizos2024survey}. Through this analysis, we aim to identify the strengths and limitations of KAN-based models in the context of data-driven weather prediction across diverse African settings.
The contribution of our work are as follows :
\begin{enumerate}
    \item We provide a comprehensive benchmarking of the recently introduced Kolmogorov-Arnold Networks (KANs) and its temporal variant TKANs against established recurrent neural network (RNN) architectures (LSTM, GRU, BiLSTM, BiGRU) and an ensemble model for forecasting key meteorological variables (precipitation, temperature at 2 meters, pressure). 
    
\item We present one of the \emph{first} benchmarks for deploying context-specific weather forecasting models in African tropical cities. Using real-world data from Abidjan and Kigali and focusing on key variables such as precipitation, temperature, and surface pressure, we generate insights that support the design of more targeted and location-specific adaptive forecasting strategies tailored to tropical African cities.

    \item We provide the \emph{first} empirical evidence of the use of TKAN to forecast short-term precipitation, temperature, and surface pressure, and demonstrate its effectiveness, particularly in two distinct regional settings (Abidjan and Kigali).

    \item We investigated alternative activation functions to the original (SiLU), namely GeLU \citep{hendrycks2023gaussianerrorlinearunits} and MiSH \citep{misra2020MiSHselfregularizednonmonotonic}, in the modeling of weather forecasts.
  
\end{enumerate}

The paper is organized as follows. In Section \ref{Lr}, we discuss the literature review on weather prediction. In Section \ref{MM}, the general approach taken in the study is introduced, including the data used and the models. Section \ref{R and D} describes the experimental results and analysis along with a discussion of different results.

\section{ Literature Review }
\label{Lr}
Weather prediction has evolved significantly over the past century, encompassing a diverse body of work ranging from physical modeling to data-driven techniques. This literature review provides a structured overview of that progression, starting with Numerical Weather Prediction (NWP), followed by statistical time-series models, advances in recurrent neural networks (RNNs), and the recent emergence of Kolmogorov–Arnold Networks (KANs) for interpretable and efficient forecasting.

The concept of weather prediction is mainly dominated by the use of numerical weather prediction (NWP) models because of their seniority.  NWP models, rooted in Vilhelm Bjerknes’ 1904 formulation of the primitive equations \citep{bjerknes2009problem}, are a set of seven differential equations (primitive equations) to model the atmosphere's state using physical laws such as mass conservation and thermodynamic laws, among others. They were later simplified by Richardson into quasi-geostrophic systems \citep{doi:10.3402/tellusa.v2i4.8607} and extended into general circulation \citep{phillips_1956,NCARGLOBALGENERALCIRCULATIONMODELOFTHEATMOSPHERE} and regional climate models \citep{kida1991new}. Despite advances, NWPs still struggle with issues such as parameterization, uncertainty of initial conditions, and costly data assimilation \citep{willis2006cleveland,pielke2006new}. To address these challenges, statistical post-processing methods such as EMOS \citep{CalibratedProbabilisticForecastingUsingEnsembleModelOutputStatisticsandMinimumCRPSEstimation} and quantile-based corrections \citep{ProbabilisticForecastsofPrecipitationinTermsofQuantilesUsingNWPModelOutput} have been developed to reduce forecast bias and improve reliability. However, two major drawbacks of NWP models were their computational complexity and their inability to effectively capture the chaotic and random dynamism of the atmosphere.  
As an alternative to NWP,  statistical models were also used for weather prediction. The first statistical method to be used without physical knowledge  was linear regression\citep{klein1959objective,billet1997use}.  Given the time-series nature of the weather and the climate data, methods based on the moving average approach, such as Autoregressive  Moving average \citep{guiot1986arma},
Autoregressive Integrated Moving average (ARIMA) \citep{sawaragi1979statistical,greenland1989climate}, Autoregressive Conditional Heteroskedasticity (ARCH)\citep{tol1997autoregressive}, Generalized Autoregressive Conditional Heteroskedasticity (GARCH)\cite{doi:10.1198/016214504000001051},  etc., have been introduced to improve the prediction performance. 
However, these methods have some limitations, including difficulty in capturing nonlinear relationships, sensitivity to missing data, and reliance on the assumption of stationarity. These challenges might result in biased or inaccurate predictions, especially when dealing with the complex and dynamic nature of meteorological data. 
In response to the limitations, deep learning, particularly recurrent neural networks (RNNs), has emerged as a powerful alternative for weather forecasting. RNNs, including their variants such as LSTM, GRU, and BiLSTM, are well suited for modeling temporal dependencies in sequential data. Several studies have demonstrated their effectiveness. For example, Gong et al. \cite{gong2024deep} proposed a CNN, LSTM hybrid model for temperature prediction, leveraging CNNs for spatial feature extraction and LSTMs for temporal modeling. Panda et al. \cite{panda2024rainfall} found BiLSTM to be superior for univariate rainfall forecasting and LSTM for multivariate scenarios, while Abbaspour et al. \cite{abbaspourclimate} and Sabat et al. \cite{sabat2022comparative} identified BiGRU as the most accurate across various regression tasks. Similarly, Guo et al. \cite{guo2024monthly} and Chhetri et al. \cite{chhetri2020deep} showed that hybrid and LSTM-based models outperform traditional approaches in predicting monthly climate variables and regional rainfall, respectively.
Despite their strengths, RNNs face challenges such as difficulty in modeling long term dependencies, limited ability to capture complex spatio-temporal interactions, and high computational cost. Moreover, issues like vanishing or exploding gradients can persist in deep architectures, even with gated units.
Kolmogorov-Arnold Networks (KANs) have recently emerged as promising alternatives to  deep learning models due to their ability to capture complex relationships through interpretable B-spline-based univariate structures, using fewer parameters. In weather forecasting, KANs have demonstrated strong potential across various tasks. Gao et al.~\citep{Gao2025} achieved a 75.33\% reduction in mean squared error when using KANs for solar radiation and 2-meter temperature forecasting in Tokyo, outperforming recurrent neural networks while maintaining interpretability. Similarly, Liu et al.~\citep{LiuKANI2025} proposed KANI, a KAN-based hypernetwork, to downscale and correct NOAA HRRR forecasts, reducing temperature and wind speed errors by 40.28\% and 67.41\%, respectively.  KAN-based models have also shown promise in hydrological contexts. Sarkar et al.~\citep{Sarkar2025} integrated KAN modules into LSTMs for flood forecasting in the Indian monsoon basin, improving peak timing and increasing Nash–Sutcliffe efficiency \citep{gupta2011typical} by 12\%. Although TKANs have not been directly applied to rainfall forecasting, their recurrent architecture, theoretically, makes them well-suited to modeling the sparsity and long-term dependencies characteristic of precipitation data. Experimentally, we have demonstrated  its effectiveness for short-term  precipitation forecasting.

Additionally, Alves et al.~\citep{Alves2024} demonstrated the efficacy of KANs in wind prediction over mountainous terrain, achieving a 48.5\% MSE reduction and yielding interpretable correction formulas. 
These findings highlight the growing relevance of KAN-based models for weather forecasting.
\section{Materials and Methods}
\label{MM}
In this section, we detail the geographical locations and characteristics of the data used in this study. The following subsections provide a detailed description of each study area and the meteorological data collected for the forecasting tasks.
\subsection{Study Area and Data }

\subsubsection{Study Area}
\begin{figure}[H]
    \centering
    \begin{minipage}{0.4\textwidth}
        \centering
        \includegraphics[height=7.5cm]{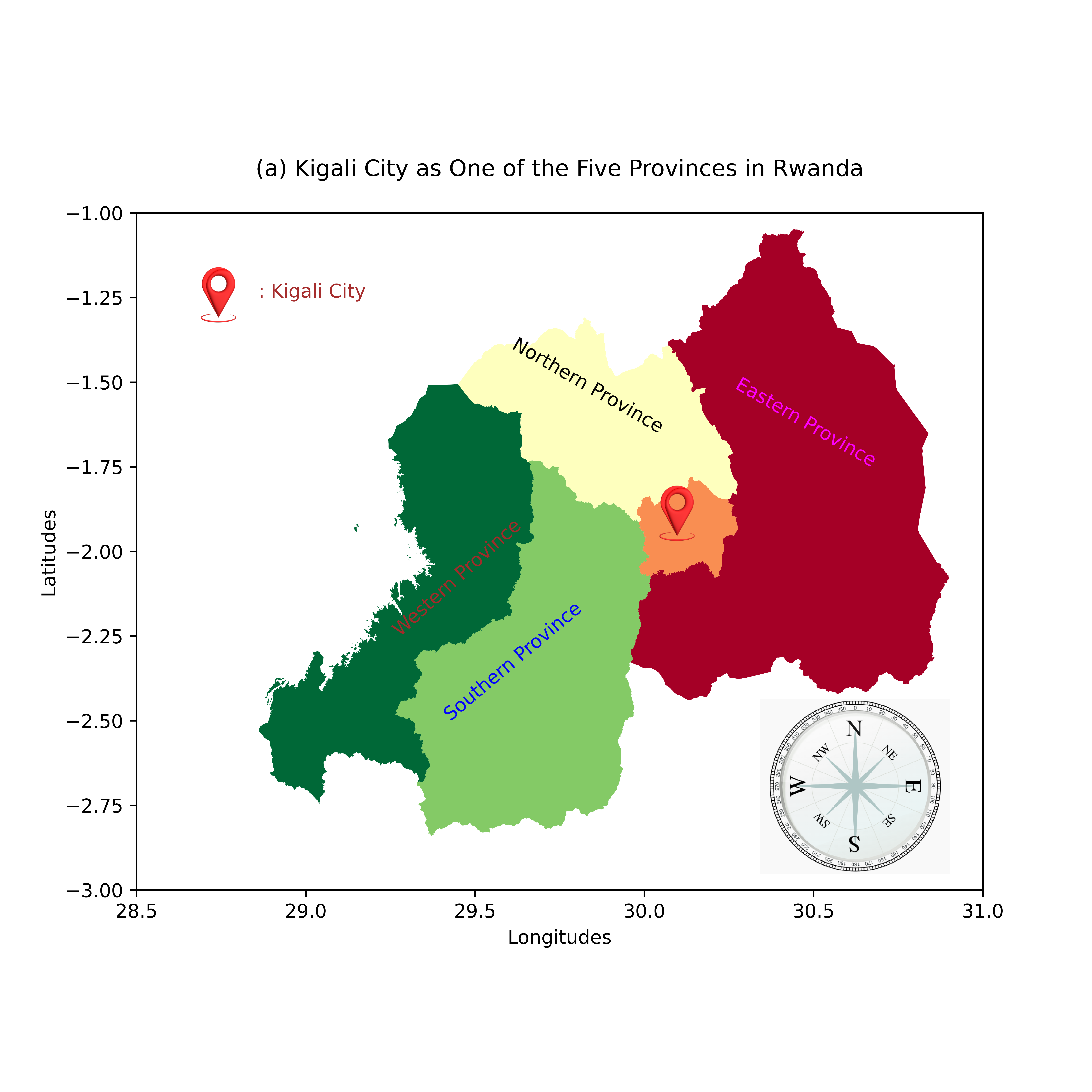}
        \caption{Kigali City}
        \label{fig:first}
    \end{minipage}
   \hfill
    \begin{minipage}{0.4\textwidth}
        \centering
        \includegraphics[height=7.5cm]{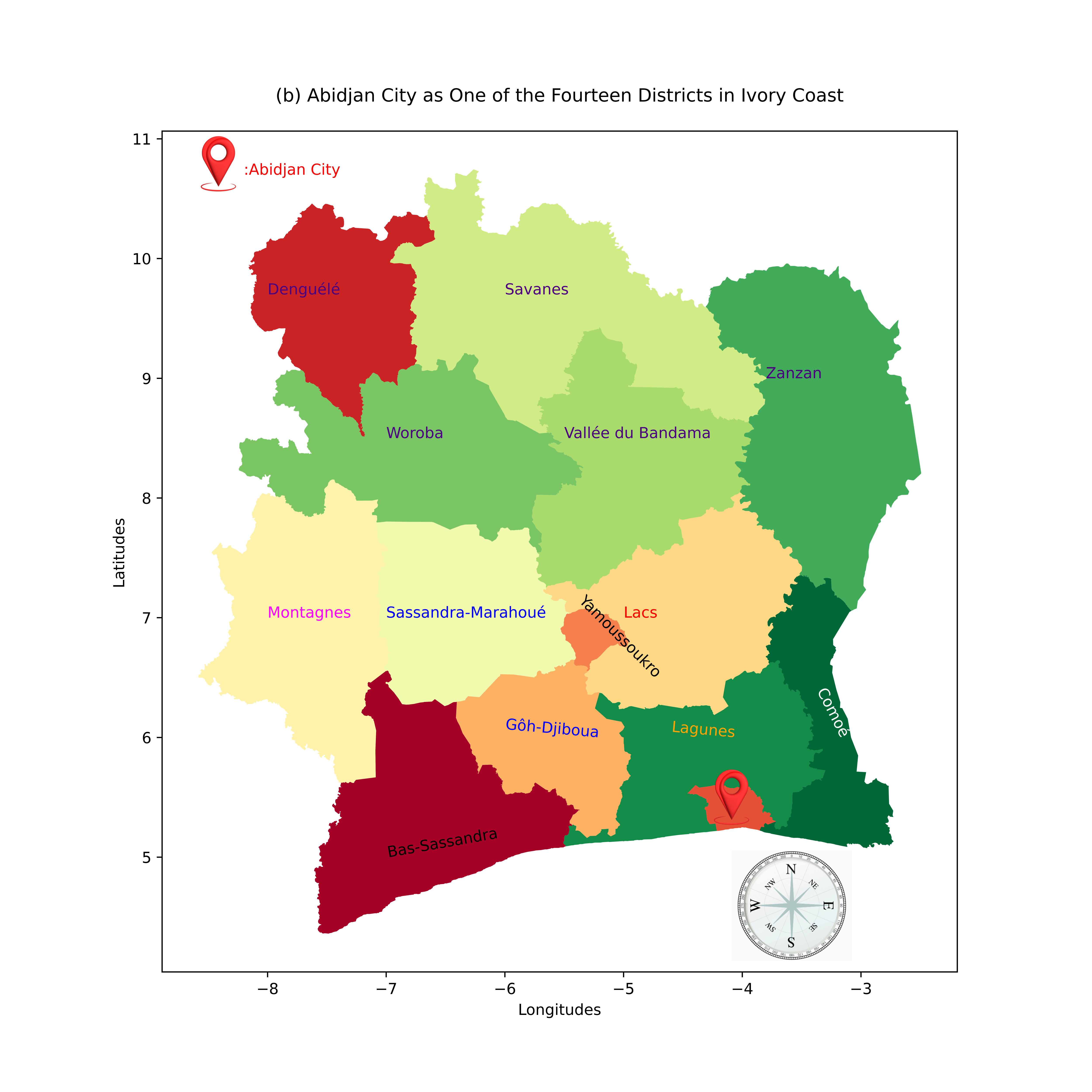}
        \caption{Abidjan City}
        \label{fig:second}
    \end{minipage}
\end{figure}

This study focuses on weather prediction in two contrasting tropical urban environments: Abidjan, Côte d'Ivoire, a low-altitude coastal city with a humid monsoon climate, and Kigali, Rwanda, a high-altitude inland city with a milder tropical savanna climate. These distinct geographical and climatic profiles enable a robust evaluation of model performance across diverse tropical settings. Study locations are shown in Figure.~\ref{fig:first} and Figure.~\ref{fig:second}. 
\subsubsection{Data}
\begin{figure}[H]
        \centering
    \includegraphics[height=3.5cm]{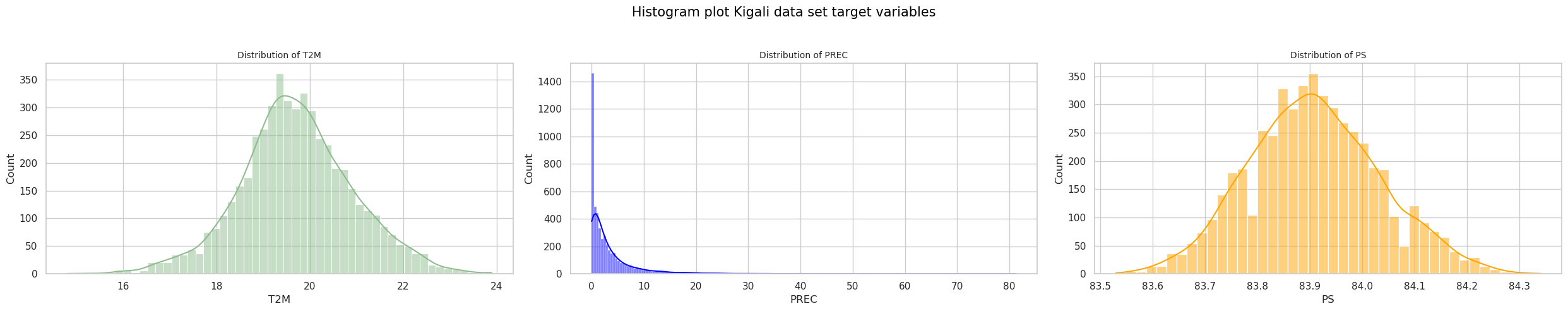} \\
    \includegraphics[height=3.5cm]{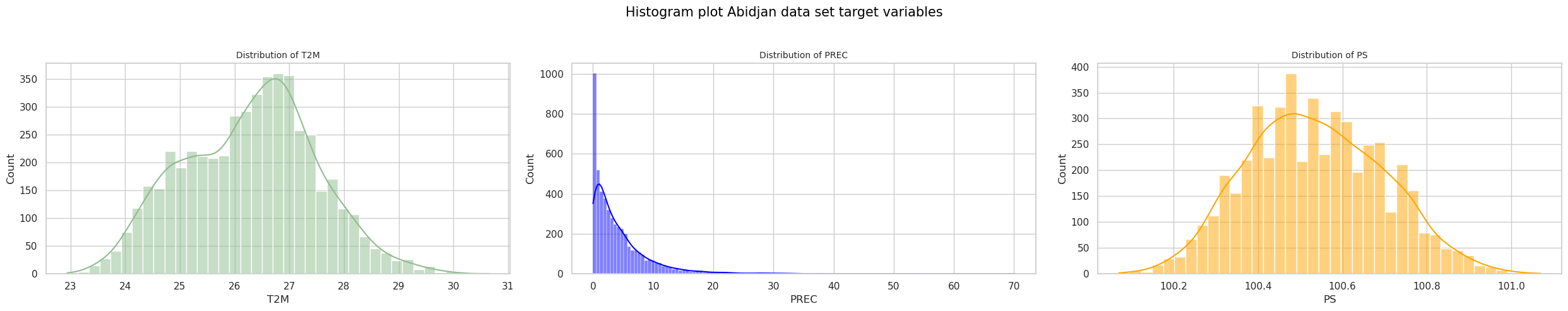}
        
         \caption{Target Variables Distribution: Row 1 is for Kigali city and row 2 for Abidjan city }
        \label{fig:kig_abj}
\end{figure}

We aim to forecast three key meteorological variables: temperature at 2 meters above sea level ($T2M$), surface pressure at 2 meters ($PS$), and precipitation ($PREC$), using datasets from Abidjan and Kigali. Figure~\ref{fig:kig_abj} displays histogram plots of the target variables for both cities. The distributions of $T2M$ and $PS$ in Kigali appear approximately Gaussian, which is  different from the case of Abidjan. In terms of PREC, Kigali seems to be a lower-precipitation zone compared to Abidjan. 
\subparagraph{Data  Information.}
The data sets\footnote{https://power.larc.nasa.gov/data-access-viewer/}, which contain 5115 observations and 10 variables, represent the behavior over a period from 01/01/2010 to 01/01/2024  of 10 weather factors in Abidjan city (Ivory Coast) and Kigali (Rwanda). The  weather factors  used are  (T2M) Temperature at 2 meters above the sea level ($^\circ C$),  (QV2M) Specific Humidity at 2 meters above the sea level  ($g/kg$), (RH2M) Relative Humidity at 2 meters above the sea level ($\%$), (PREC)  Precipitation  (mm/day), (PS) Surface Pressure (kPa), (SWDWN) All Sky Surface Shortwave Downward Irradiance ($kW{-}hr/m^2/day$), (CSWDWN) Clear Sky Surface Shortwave Downward Irradiance ($kW{-}hr/m^2/day$), (LWDWN) All Sky Surface Longwave Downward Irradiance ($W/m^2$), (T2MDEW) Dew/Frost Point at 2 meters (C),(T2MWET) Wet Bulb Temperature at 2 meters (C).				

\subparagraph{Data Set Manipulation.}

The dataset was split into 72\% for training, 8\% for validation, and 20\% for testing. For preprocessing, we applied the \texttt{MinMaxScaler} with a feature range of $[0, 1]$ for the \texttt{T2M} and \texttt{PS} variables, and $[-1, 1]$ for \texttt{PREC} when using recurrent neural network-based models. For the other models, all variables were scaled to the $[0, 1]$ range. The spline-based models specifically required input scaling to $[0, 1]$ to ensure compatibility with the predefined grid of the B-spline basis functions. We used a batch size of 64 for all experiments and constructed the sequential datasets using a sliding window approach, considering 14 days of observations to predict the next day.

\subsection{Methods}
For the sake of completeness, we briefly present the models used in this study and refer to the references contained therein for more details.
\subsubsection{ Long Short Term Memory (LSTM)}
Long-Short-Term Memory (LSTM) networks\cite{artr} are a type of recurrent neural network(RNN) designed to capture long-term dependencies in sequential data. Each LSTM cell maintains a memory state regulated by three gates: the forget gate, which determines which information to discard; the input gate, which controls what new information to store, and the output gate, which decides the information to store. These gates enable the model to forget, retain or update information effectively over time.
At each time step, the LSTM processes the current input $x_t$, the previous hidden state $h_{t-1}$. and the previous cell state $C_{t-1}$ to produce a new hidden state $h_t$, a new memory state $C_{t}$, and an output $\hat{y}_t$. The computations are governed by nonlinear functions and learnable parameters that adjust during training to model temporal dependencies in data.

\subsubsection{Gated recurrent units (GRU)}
Gated Recurrent Units (GRU) \citep{cho-etal-2014-properties} are simplified variant of LSTM that are efficiently capture dependencies across various time scales. Unlike LSTM, GRU merges the cell and hidden states, using fewer gates to control the flow of information. The core components include the update gate $z_t$, which balances the contribution of the previous hidden state and new candidate information; the reset gate $r_t$, which controls how much past information to forget; and the candidate activation $\tilde{h}_t$, which forms the basis of new hidden state $h_t$.
These gates enable GRU to maintain performance comparable to LSTM while offering reduced computational complexity due to their streamlined structure.

\subsubsection{Bidirectional Recurrent Neural Network-Based Models}
The bidirectional framework considers both information transmission: utilizing the past values to enhance the upcoming values through a layer called the forward layer ($L^{for}$), as well as using the backward layer($L^{back}$) that processes the next values to rectify the previous values. After forward and backward processing, the results of both layers are aggregated to constitute the output of the algorithm.

This principle was used with  LSTM and GRU  layers to developed the so-called Bi-LSTM and Bi-GRU \citep{bin2018describing,Su_2019,cheng2019data}.



\subsubsection{ Ensemble Model}
Let \( \{f^i_{\theta^i}\}_{i=1}^4 \) be four recurrent models (LSTM, GRU, BiLSTM, BiGRU) parameterized by \( \theta^i \). Considering the multivariate time series data $X=\{x_1,\dots,x_T\}$ where each $x_t \in \mathbb{R}^d$, we  assume a target variable $y_t\in x_t$,
the ensemble model used can be described as
\begin{equation}
    F(y_T)=\sum_{i=1}^{4} C_{\theta^i}f^i_{\theta^i},(x_1,\cdots,x_{T-1})
\end{equation}
where $C_{\theta^i}$ are  trainable  coefficients between [0,1] (we applied the softmax function to restrict their values between [0,1]).

\subsubsection{Kolmogorov Arnold Network (KAN)}
The Kolmogorov-Arnold representation theorem \cite{Arnold2009} establishes that every continuous function of multiple variables can be represented by a sum of continuous univariate functions and the binary operation of addition. More specifically, for a smooth $f:[0,1]^m\to\mathbb{R}$,
\begin{equation}\label{eq:KART}
    f(x) = f(x_1,\dots,x_n)=\sum_{q=1}^{2n+1} \Phi_q\left(\sum_{p=1}^n\phi_{q,p}(x_p)\right),
\end{equation}
where $\phi_{q,p}:[0,1]\to\mathbb{R}$ are  the univariate functions (B-splines in this case) $\Phi_q:\mathbb{R}\to\mathbb{R}$  are continuous functions.

Kolmogorov-Arnolds networks (KANs) \cite{liu2024kan} leverage the Kolmogorov-Arnold representation theorem  by replacing traditional linear function in neural networks with spline-parametersized univariate functions. Unlike conventional Multi-Layer Perceptrons (MLPs) that utilize fixed activation functions at their nodes(neurons), KANs employ adaptive, learnable activation functions on the edges connecting nodes. These edge functions are parametrized as B-spline curves, which dynamically adjust during training to better model the underlying data patterns. This unique architecture enables KANs to effectively capture complex nonlinear relationships within the data. Formally, a KAN layer can be defined as $\Phi = \{\phi_{q,p}\}$, where $p = 1, 2, \dots, n_{in}$ and $q = 1, 2, \dots, n_{out}$, with $\phi_{q,p}$ being parametrized functions with learnable parameters. This structure allows KANs to capture complex nonlinear relationships within the data more effectively than traditional Multi-Layer Perceptrons (MLPs).
To enhance the modeling power of KANs, deeper architectures have been created by stacking multiple KAN layers \cite{liu2024kan}. This composition of layers allows the network to learn more complex functions. The architecture of a deeper KAN can be expressed as:
\begin{equation}
\text{KAN}(x) = (\Phi_{L-1} \circ \Phi_{L-2} \circ \dots \circ \Phi_{0})(x).  
\end{equation} where  each $\Phi_{l}$ denotes a KAN layer.
In this experiment, we use the layer formula  provided by  \cite{Arnold2009}, defined as follows \begin{equation}
    \Phi_{l}=W_bb(x)+W_sSpline(x)
    \label{silu}
\end{equation}
where $b(x)$ is the SiLU activation function \citep{elfwing2017sigmoidweightedlinearunitsneural} and $Spline(x)=\sum_{i=1}^{N}C_iB_i(x)$ where all  $B_i$ are B-spline and  $C_i$ are trainable parameters; $N$ is the number of spline.
Increasing network depth enables the capture of more complex data patterns and dependencies, with each layer $l$ transforming the input $x$ through learnable functions $\phi_{q,p}$, resulting in a highly adaptable and powerful model.

\subsubsection{Temporal Kolmogorov-Arnold Networks}
Temporal Kolmogorov-Arnold Networks (TKAN) \cite{genet2024tkan} is a neural network architecture that merges Kolmogorov-Arnold Networks (KANs) with memory management techniques taken from LSTMs. It  incorporates LSTM-like gates to manage long-term dependencies coupled with the KAN layer  output to handle  sequential data.  At every time step \( t \), the input \( x_t \) alongside the previous memory state \( \tilde{h}_{l,t-1} \) ($l\in\{1,\dots,L\}$ represents the number of KAN layers) is merged to determine the  current sub layer hidden $\tilde{h}_{l,t}$ sate (temporal information provided by the B-spline  sublayers)  as follows:
\begin{equation}
\tilde{h}_{l,t} = W_{hh} \tilde{h}_{l,t-1} + W_{hz} \tilde{o}_t
\end{equation}
where 
\begin{equation}
\tilde{o}_t = \phi_l(s_{l,t}),  ~~\text{with}, ~~ s_{l,t} = W_{l,\tilde{x}} x_t + W_{l,\tilde{h}} \tilde{h}_{l,t-1}
\label{kan_sub}
\end{equation}
This architecture incorporates short-term memory, enabling the sub-layer  to efficiently understand dependencies in time series data. TKAN utilizes LSTM-like gating mechanisms to manage long-term memory. 
Let $x_t$ be the input vector of dimension d. This unit uses several internal vectors and gates to manage information flow \cite{genet2024tkan}. The forget gate with activation vector $f_t$ given by
\begin{equation}
f_t = \sigma(W_f x_t + U_f h_{t-1} + b_f),
\end{equation}
decides what information to forget from the previous state. The input gate, with activation vector denoted $i_t$, defined as follow
\begin{equation}
i_t = \sigma(W_i x_t + U_i h_{t-1} + b_i),
\end{equation}
defines which information to insert into memory.
The  cell state $c_t$, responsible to provide timely raw temporal information to the memory  is updated using
\begin{equation}
c_t = f_t \odot c_{t-1} + i_t \odot \tilde{c}_t.
\end{equation}
The Candidate memory is given by 
\begin{equation}
\tilde{c}_t = \tanh(W_c x_t + U_c h_{t-1} + b_c).
\end{equation}
The KAN-based short-term information gotten from eq. \eqref{kan_sub} is then concatenated into a cell $r_t $ which contains temporal information from diverse KAN layers
\begin{equation}
\label{r_t}
r_t = \text{Concat}[\phi_1(s_{1,t}), \phi_2(s_{2,t}), \ldots, \phi_L(s_{L,t})]
\end{equation}
Finally, the output gate and the hidden state are  computed by
\begin{equation}
o_t = \sigma(W_o r_t + b_o)
\end{equation}
\begin{equation}
h_t = o_t \odot \tanh(c_t)
\end{equation}
with \( \odot \)  and \( \sigma \), $r_t$  represent element-wise multiplication, the sigmoid activation function and the concatenation of the output of multiple KAN Layers respectively. For our experiments, in Eq. \eqref{r_t}, we used $L=1 ~~ \text{and} ~~ L=5 $ for TKAN and TKAN(5 Sub-layer) respectively. In addition to the SiLU activation function, we  also replaced the  b(x) function in eq. \eqref{silu} by the GeLU \citep{hendrycks2023gaussianerrorlinearunits} and MiSH \citep{misra2020MiSHselfregularizednonmonotonic} activations in the case of TKAN and denote them respectively GeLU TKAN and MiSH TKAN.



\section{ Experimental Results  and Discussion} 
\label{R and D}
This section provides a detailed discussion of the results obtained following our analysis and methodology. All the implementation in this work was done using the \textbf{Pytorch} \citep{paszke2019pytorch} Library. Note that we implemented the KAN function using the default parameters (As provided in the initial paper\citep{liu2024kan}). We trained  all our models using the Adam optimizer with learning rate $lr=0.001$. We used layer dropout (dropout rate $p=0.2$) for the deep RNN-based models. The code used in this study can be found here \footnote{ \url{https://github.com/AngeClementAkazan/Localized-Weather-Prediction-Using-KAN-and-DeepRNNs}}. 
\begin{table}[ht!]
\centering
\resizebox{0.8\textwidth}{!}{\begin{tabular}{lllc lrrrrr}
\toprule
Dataset &  Model &  MSE $\downarrow$ & RMSE $\downarrow$ & MAE $\downarrow$& R² $\uparrow$&  MAPE (\%) $\downarrow$ \\
\midrule
\textbf{Abidjan data} &  LSTM &  35.9654  &	5.9971 &	3.3102 &	0.2625 &	347.9945 \\
&  BiLSTM &   36.5676 &	6.0471 &	3.1866	& 0.2501 &	359.4971 \\
&  GRU &   35.6789 &	5.9732 &	3.3231 &	0.2683 &	365.2702   \\
&  BiGRU &   37.0915 &	6.0903 &	3.1288 &	0.2394 &	261.9967   \\ 
&  Ens. Mod &   35.9182	& 5.9932 &	\textbf{3.0776} &	0.2634 &	292.3611\\
&  KAN &  \textbf{35.211} &	\textbf{5.9339}	& 3.3438	& \textbf{0.2698}	& \textbf{231.5354}    \\
&  TKAN (5 Sub-layers) &  37.868 &  6.1537 &  3.2478 &  0.2235 &  296.3412   \\
&  TKAN  & 36.2615 & 6.0218 & 3.2198 & 0.2564 & 298.0565 \\
&MiSH TKAN& 35.882 & 5.9902 & 3.2386 & 0.2642 & 309.2074 \\
&GeLU TKAN & 36.6378 & 6.0529 & 3.3254 & 0.2487 & 346.1602 \\
\hline  
\textbf{Kigali data}  &  LSTM &   29.4953 &	5.431 &	3.0459 &	0.2145	& 842.8203   \\
&  BiLSTM &  29.0966 &	5.3941 &	2.9134 &	0.2251 &	714.2877   \\
&  GRU &  29.4448 &	5.4263 &	3.1399 &	0.2158	& 902.884   \\
&  BiGRU &  29.8353	& 5.4622 &	3.0244	& 0.2055 &	823.4082  \\ 
&  Ens. Mod & 30.23 & 	5.4982	& 2.9092	& 0.1949 &	651.7455 \\
&  KAN &   27.9707	& 5.2887 &	2.8447 &	\textbf{0.2463} &	\textbf{428.0795 }  \\

&  TKAN (5 Sub-layers) &  23.5482 &  4.8526 &  \textbf{2.4067 }&  0.1568 &  539.4666     \\
&  TKAN &\textbf{21.4602} & \textbf{4.6325} & 2.5796 & 0.2315 & 743.6441\\
&MiSH TKAN &21.5305 & 4.6401 & 2.5809 & 0.229 & 734.7372\\
&GeLU TKAN & 22.3205 & 4.7245 & 2.6161 & 0.2007 & 828.9953 \\
\bottomrule
\end{tabular}
}
\caption{Performance metrics for different models  across datasets for precipitation (Best in \textbf{Bold})}
\label{tab:metrics1}
\end{table}
\newline \newline
In table \eqref{tab:metrics1}, the precipitation forecasting  evaluation across both datasets (Abidjan and Kigali) highlights distinct model strengths. On the Abidjan dataset, the KAN model demonstrated the most best  performance, achieving the lowest MSE ($35.21$ mm\textsuperscript{2}) and RMSE ($5.93$ mm), as well as the highest $R^2$ value ($0.2698$), indicating both precision and strong variance explanation. However, its MAE was slightly higher compared to the Ensemble model ($3.08$ mm) and BiGRU ($3.13$ mm), suggesting KAN may exhibit a few larger absolute deviations. The notably high MAPE values across all models are a result of the inherently low precipitation levels in both cities, which inflate relative error calculations.
In terms of MAPE, KAN again led ($231.54\%$), confirming its robustness against relative errors under low precipitation conditions. The poorest performance was observed with the TKAN (5 Sub-layers) model, which recorded the highest MSE and the lowest $R^2$.  For the Kigali dataset, TKAN-based models achieved, respectively, the best MSE, RMSE and MAE, demonstrating the effectiveness of their importance in this context. Nonetheless, KAN  achieved the highest $R^2$ ($0.2463$) and the lowest MAPE ($428.08\%$), indicating superior generalization and robustness.
 Notably, the TKAN variant that used the MiSH activation function (MiSH TKAN) demonstrated more consistent performance than its GeLU and SiLU-based counterpart.
 \begin{figure}[H]
    \centering
    \begin{subfigure}{0.4\textwidth}
        \centering
        \includegraphics[height=4cm]{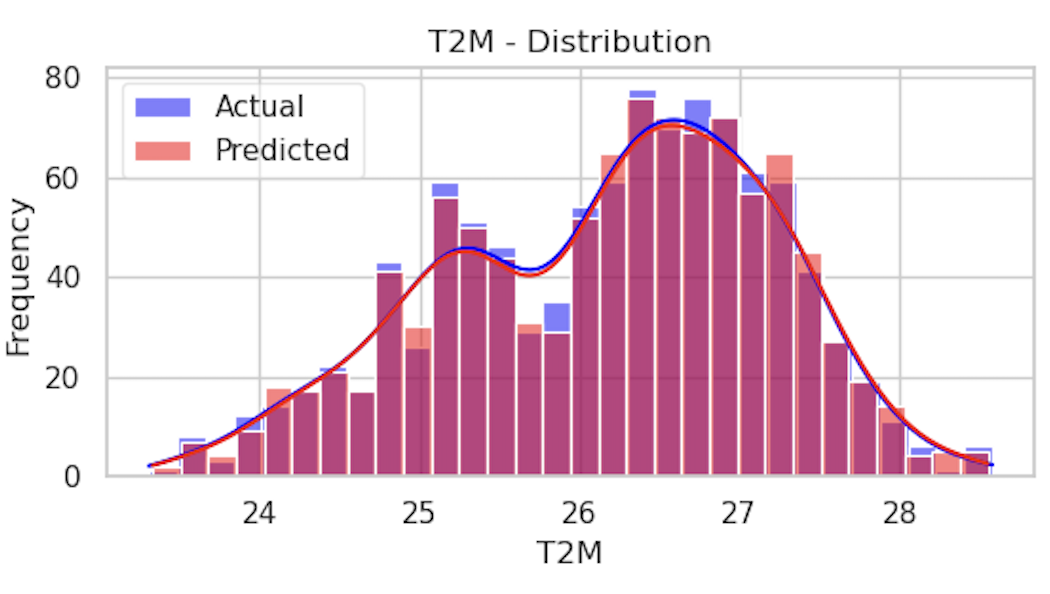}
        \caption{Kolmogorov-Arnold Networks}
        \label{fig:kanpred_}
    \end{subfigure}
   \hfill
    \begin{subfigure}{0.4\textwidth}
        \centering
        \includegraphics[height=4cm]{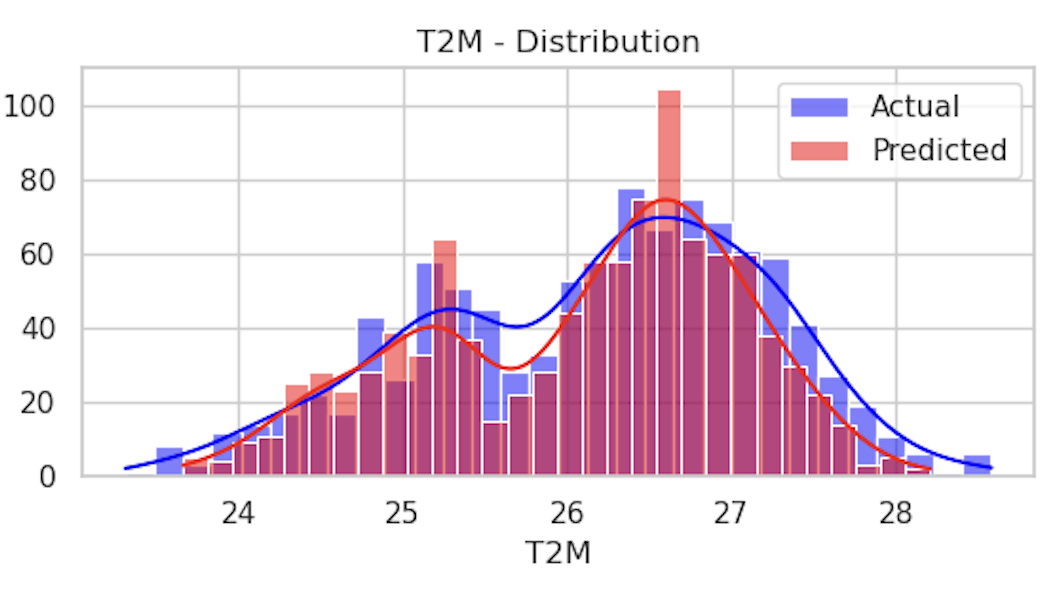}
        \caption{BiGRU}
        \label{fig:bigru_pred}
    \end{subfigure}
    \caption{Comparison of predicted vs. actual temperature values in Abidjan using the best-performing KAN-based model and the top-performing deep RNN model}
     \label{fig:bothcities}
\end{figure}

\begin{figure}[H]
    \centering
    \begin{subfigure}{0.4\textwidth}
        \centering
        \includegraphics[height=4cm]{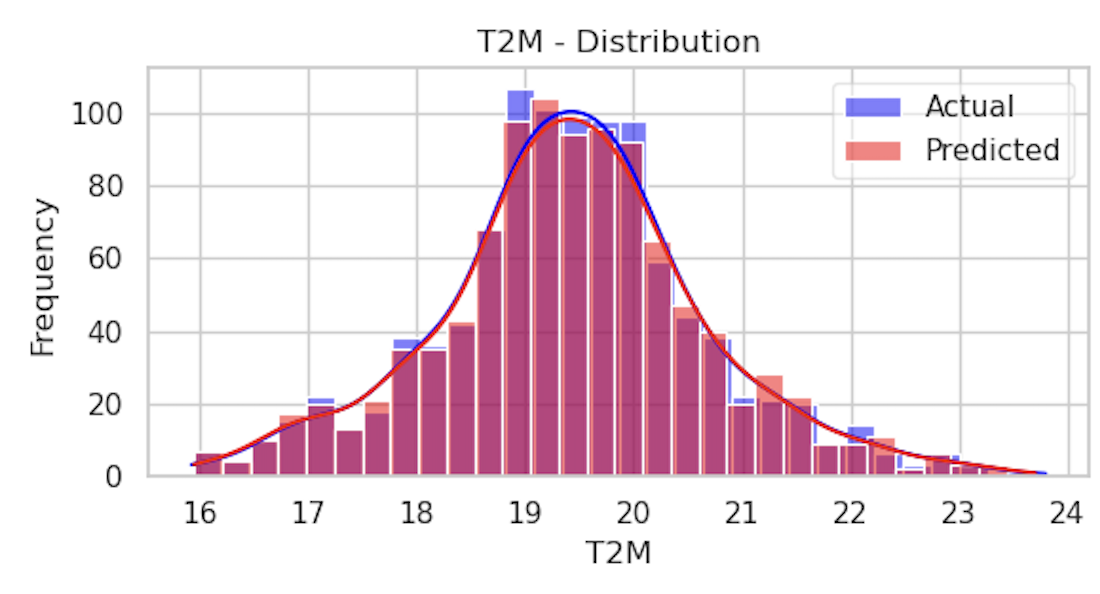}
        \caption{Kolmogorov-Arnold Networks }
        \label{fig:kanpred_2}
    \end{subfigure}
   \hfill
    \begin{subfigure}{0.4\textwidth}
        \centering
        \includegraphics[height=4cm]{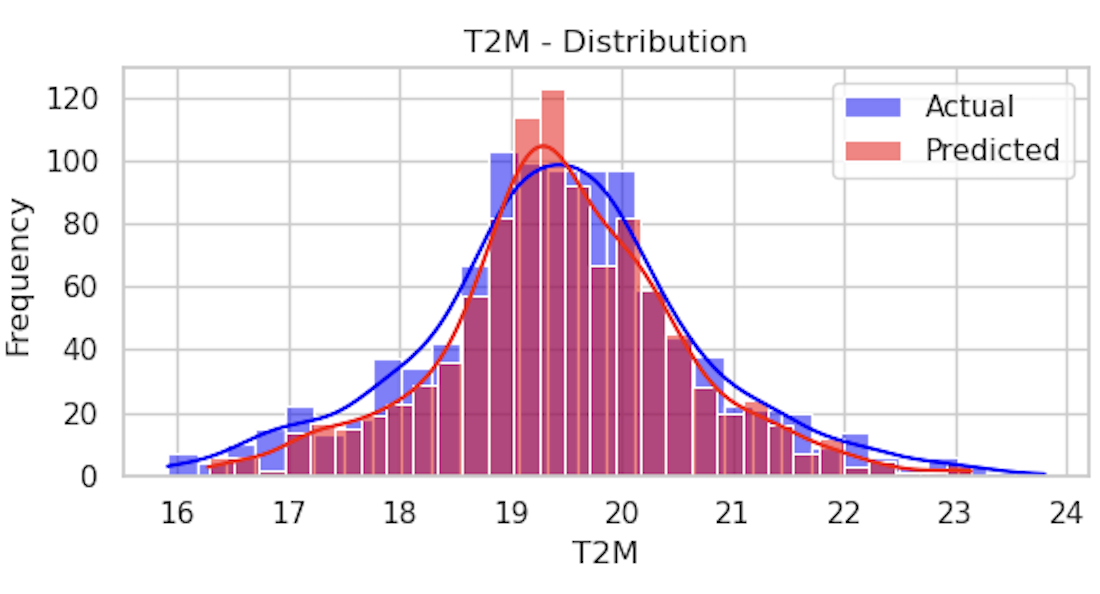}
        \caption{BiGRU}
        \label{fig:bigru_pred_2}
    \end{subfigure}
    \caption{Comparison of predicted vs. actual temperature values in Kigali using the best-performing KAN-based model and the top-performing deep RNN model}
    \label{fig:bothcities_1}
\end{figure}

        
\begin{table}[ht!]
\centering
\resizebox{0.8\textwidth}{!}{\begin{tabular}{lllc lrrrrr}
\toprule
Dataset &  Model &  MSE $\downarrow$ & RMSE $\downarrow$  & MAE $\downarrow$ & R² $\uparrow$ &  MAPE $\downarrow$ \\
\midrule
\textbf{Abidjan data} &  LSTM & 7.2616	& 2.6947 &	2.052 &	0.8349 &	3.2282  \\
&  BiLSTM &  6.3758	& 2.525	& 1.8982 &	0.855 &	2.9552  \\
&  GRU &  7.4861 &	2.7361 &	2.0823 &	0.8298 &	3.2665
  \\
&  BiGRU &   6.222 &	2.4944 &	1.9174 &	0.8585 &	2.9405 \\ 
&  Ens. Mod &   6.271 &	2.5042 &	1.9276 &	0.8574 &	2.9558 \\
&   KAN &  \textbf{0.0014}	& \textbf{0.0377}&	\textbf{0.0205} &	\textbf{0.9986}	& \textbf{0.0781} \\
&  TKAN (5 Sub-layers)& 0.1491 &	0.3861 &	0.2903	&0.8592	& 1.1112
\\
&  TKAN  & 0.1462 & 0.3824 & 0.2897 & 0.8619 & 1.1066 \\
& MiSH TKAN &0.1586 & 0.3982 & 0.303 & 0.8502 & 1.1615\\
&GeLU TKAN&0.1509 & 0.3885 & 0.2947 & 0.8574 & 1.1284\\

\hline 
\textbf{Kigali data}  &  LSTM &   12.4869 &	3.5337	& 2.7118	& 0.7416 &	4.067 \\
&  BiLSTM &   11.8499 &	3.4424	& 2.6935 & 0.7548 &	3.9444 \\
&  GRU &   12.0403 &	3.4699 &	2.675 &	0.7509  & 3.9882  \\
&  BiGRU &   11.8098 &	3.4365 & 	2.6424 &	0.7556	&3.9065  \\ 
&  Ens. Mod &  11.9369	& 3.455 &	2.666 &	0.753 &	3.9414 \\
& KAN &  \textbf{0.0003}	& \textbf{0.0177} &	\textbf{0.0138} &	\textbf{0.9998}	& \textbf{0.0714}    \\
&  TKAN (5 Sub-layers) & 0.2808 &  0.5299 &  0.408 &  0.7508 &  1.5258     \\
&  TKAN  &0.2726 & 0.5221 & 0.4027 & 0.758 & 1.5036  \\
&MiSH TKAN&0.272 & 0.5215 & 0.4045 & 0.7585 & 1.5095\\
&GeLU TKAN&0.2686 & 0.5182 & 0.399 & 0.7616 & 1.488 \\
\bottomrule
    
\end{tabular}
}
\caption{Performance metrics for different models  across datasets for temperature at 2 meters (Best in \textbf{Bold})}
\label{tab:metrics2}
\end{table}

Table \eqref{tab:metrics2}, fig.\eqref{fig:bothcities} and fig.\eqref{fig:bothcities_1} showed that the Kolmogorov-Arnolds Networks (KAN) model demonstrated a largely superior performance compared to all other evaluated models (LSTM, BiLSTM, GRU, BiGRU, Ensemble model and TKAN variants ) while predicting temperature (unit$^\circ C$) on both Abidjan and Kigali datasets. KAN achieved significantly higher accuracy across all metrics, indicating a good fit. The other models showed considerably poorer performance, although they exhibited expected trends among themselves (e.g., bidirectional-based RNNs slightly outperforming unidirectional ones).  The TKANs with SiLU and GeLU activations typically showed slightly better performance than  using MiSH TKAN. The TKANs also marginally (in Kigali) outperform the  TKAN (5 Sub-layers). Despite the performance with TKAN using SiLU and GeLU activations functions, all the TKAN models were significantly less accurate than the KAN model. This result strongly suggests that KAN is the most effective temperature model. \footnote{Visualizations of the predicted versus actual distributions, including line plots and histograms, are provided in Appendix, (\autoref{appendix}).}



\begin{table}[ht!]
\centering
\resizebox{0.8\textwidth}{!}{\begin{tabular}{lllc lrrrrr}
\toprule
Dataset &  Model &  MSE $\downarrow$ & RMSE $\downarrow$  & MAE $\downarrow$ & R² $\uparrow$ &  MAPE $\downarrow$\\
\midrule
\textbf{Abidjan data} &  LSTM & \textbf{0.0036} &	\textbf{0.0602} &	\textbf{0.0483} &	\textbf{0.8539} &	0.0481  \\
&  BiLSTM &   0.0037 & 0.0606	& 0.0481 & 0.8521 &	\textbf{0.0478} \\
&  GRU &   0.0037 &	0.0611 &	0.0492 &	0.8496 &	0.0489  \\
&  BiGRU &   0.0037	& 0.0606 &	0.0484 &	0.852 &	0.0481 \\ 
&  Ens. Mod &   0.0036 &	0.0603 &	0.0481 &	0.8534 &	0.0479 \\
& KAN &   0.012	& 0.1095 &	0.0882 &	0.5171 &	0.0877 \\

&  TKAN(5 Sub-layers) & 0.0038 &  0.0615 &  0.0494 &  0.8479 &  0.0491 \\
&  TKAN  &   0.004 &  0.0632 &  0.0509 &  0.8394 &  0.0506     \\
&MiSH TKAN&0.0039 & 0.0625 & 0.0501 & 0.8425 & 0.0498 \\
&GeLU TKAN& 0.0038 & 0.0618 & 0.0494 & 0.8464 & 0.0492\\
\hline 
\textbf{Kigali data}  &  LSTM & \textbf{0.0024}	& \textbf{0.0491} &	\textbf{0.0385}	& \textbf{0.8593} &  \textbf{0.0459} \\
&  BiLSTM &  0.0025 &	0.0498 &	0.0393 &	0.8552 &	0.0468  \\
&  GRU &  0.0025 &	0.0496 &	0.039 &	0.8564	& 0.0464  \\
&  BiGRU &  0.0026 &	0.0509 &	0.0401	& 0.8484 &	0.0478  \\ 
&  Ens. Mod & 0.0025 &	0.0501 &	0.0395 &	0.8537 &	0.0471 \\
&  KAN &    0.0087 &	0.0934	& 0.0746 &	0.4966	& 0.0889 \\
&  TKAN (5 Sub-layers) &  0.0039 &  0.0621 &  0.0491 &  0.8521 &  0.0488 \\
&  TKAN  &0.0039& 0.0627 & 0.0495 & 0.8492 & 0.0492    \\
&MiSH TKAN&0.0038 & 0.0619 & 0.0492 & 0.8531 & 0.0489\\
&GeLU TKAN&0.0039 & 0.0623 & 0.0491 & 0.8516 & 0.0488\\
\bottomrule

\end{tabular}
}
\caption{Performance metrics for different models  across datasets for pressure (Best in \textbf{Bold})}
\label{tab:metrics3}
\end{table}
Table~\ref{tab:metrics3} summarizes the performance metrics  for the atmospheric pressure prediction (units in kPa), evaluated across  the Abidjan and Kigali datasets. Standard recurrent neural networks (LSTM, GRU), their bidirectional variants (BiLSTM, BiGRU), and the Ensemble model exhibited stronger and more consistent performance on both datasets. 
In  contrast, KAN showed notably lower predictive accuracy, reflected by significantly poor regression errors on both datasets (Abidjan and Kigali).
Introducing temporal structures into KAN (TKAN and TKAN with 5 Sub-layers) significantly improved predictions, nearly matching deep RNN-based model performances on both datasets (especially for Abidjan data). On a side note, the inception of new activations (GeLU and MiSH) consistently improve TKAN.


\section{Conclusion and Future Work}
\label{conc}
In this study, we addressed the challenging task of short-term, localized weather forecasting for three key meteorological variables, precipitation, temperature at 2 meters, and surface pressure, in two African capital cities: Abidjan (Côte d'Ivoire) and Kigali (Rwanda). We conducted a comparative evaluation of well-known recurrent neural networks (LSTM, GRU, BiLSTM, and BiGRU), a standard ensemble model, and a set of emerging models based on Kolmogorov–Arnold representations. These included the original Kolmogorov–Arnold Network (KAN), its temporal extension  (TKAN), two customized versions of TKAN that replaced the standard SiLU activation function with either GeLU or MiSH, resulting in the GeLU TKAN and MiSH TKAN variants, and  TKAN (with 5 Sub-layers). Using real daily meteorological data from Abidjan and Kigali, our findings revealed that classical RNNs performed well in forecasting surface pressure but struggled to generalize across weather variables and locations. Their performance dropped noticeably when predicting temperature and precipitation, highlighting limitations in adapting to different climatic conditions. In contrast, KAN-based architectures excelled, particularly in temperature forecasting, achieving exceptionally low error metrics (MSE, MAE, RMSE, MAPE) and $R^2$ values exceeding 0.99 in both cities. They also delivered competitive results for precipitation. However,  despite their strength, the same models were less effective at predicting pressure, suggesting difficulty in modeling this variable’s temporal dynamics across diverse settings. Among the KAN variants, the introduction of GeLU and especially MiSH activation functions consistently improved the performance of TKAN. Overall, the results highlight the potential of KANs and related architectures as flexible and powerful alternatives to traditional RNNs in climate modeling. Notably, the integration of alternative activation functions, such as GeLU and MiSH, demonstrated significant improvements in TKAN-based models. 

Yet, the findings also underscore the need for more robust variants beyond TKAN, models capable of consistently performing across different weather variables and diverse climatic settings. One promising direction for future work is to refine KAN-based architectures through systematic hyperparameter tuning, such as grid or random search. Additionally, modifying the temporal component of TKAN, by exploring alternative kernel structures, adjusting the spline order, number or placement of B-splines, or integrating attention mechanisms, could further enhance its adaptability and performance across varying meteorological contexts.

\section*{Acknowledgments}  We extend our sincere gratitude to Athanase Hafashimana for the insightful discussions on the subject and for providing the images of the study area, which greatly contributed to the progress of this work.

\newpage
\bibliography{main}
\bibliographystyle{plain}
\appendix
\section*{Appendix}
\label{appendix}
\addcontentsline{toc}{section}{Appendix}

This appendix offers a visual exploration of the forecasting outcomes, presenting a series of images that showcase representative predictions from and Kolmogorov-Arnoal Networks based models and the different Deep Recurrent Neural Networks  models alongside the corresponding ground truth data for temperature at 2 meters, precipitation, and atmospheric pressure in Abidjan and Kigali.

\begin{figure}[H]
        \centering
        \includegraphics[height=6.5cm]{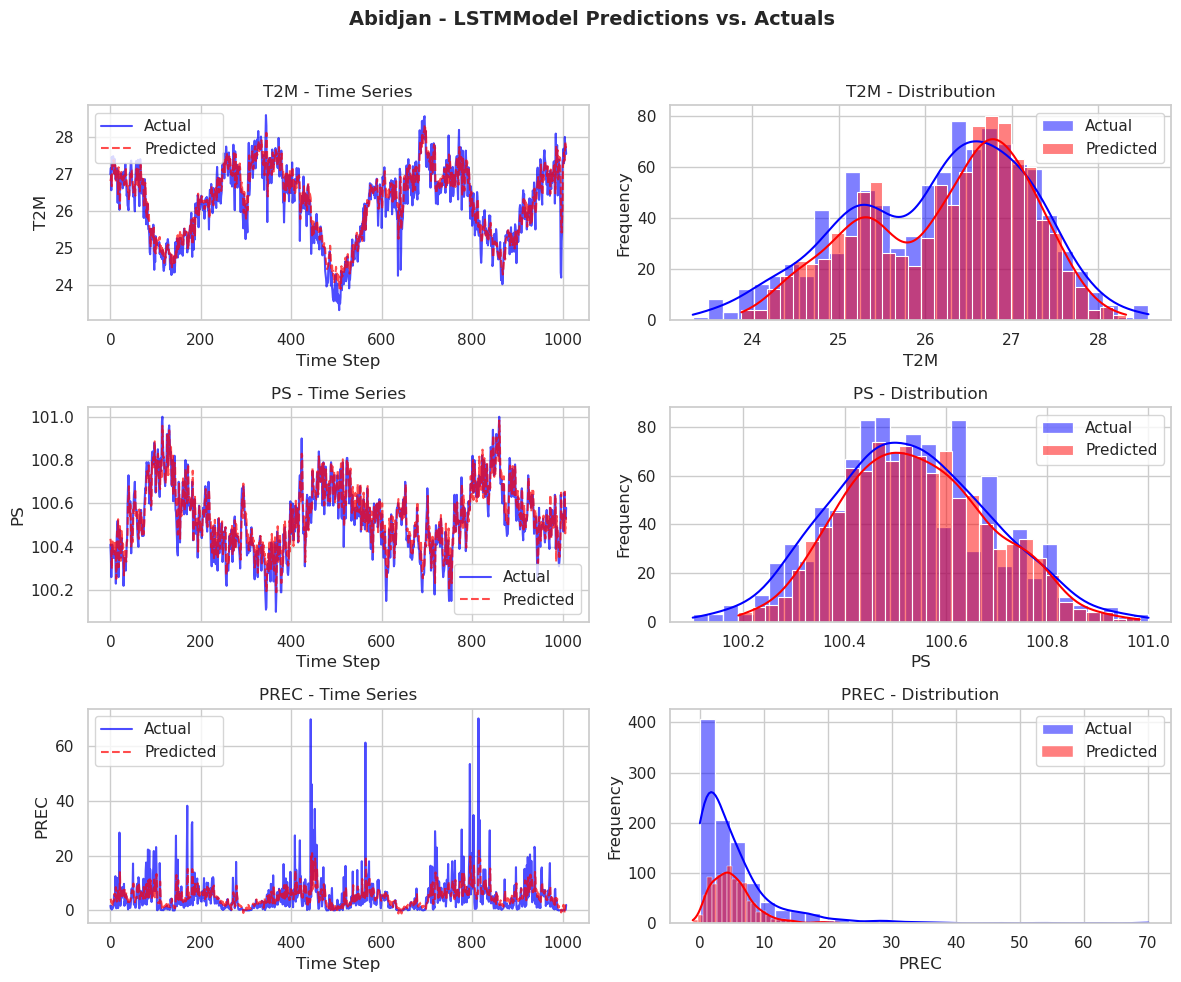} \includegraphics[height=6.5cm]{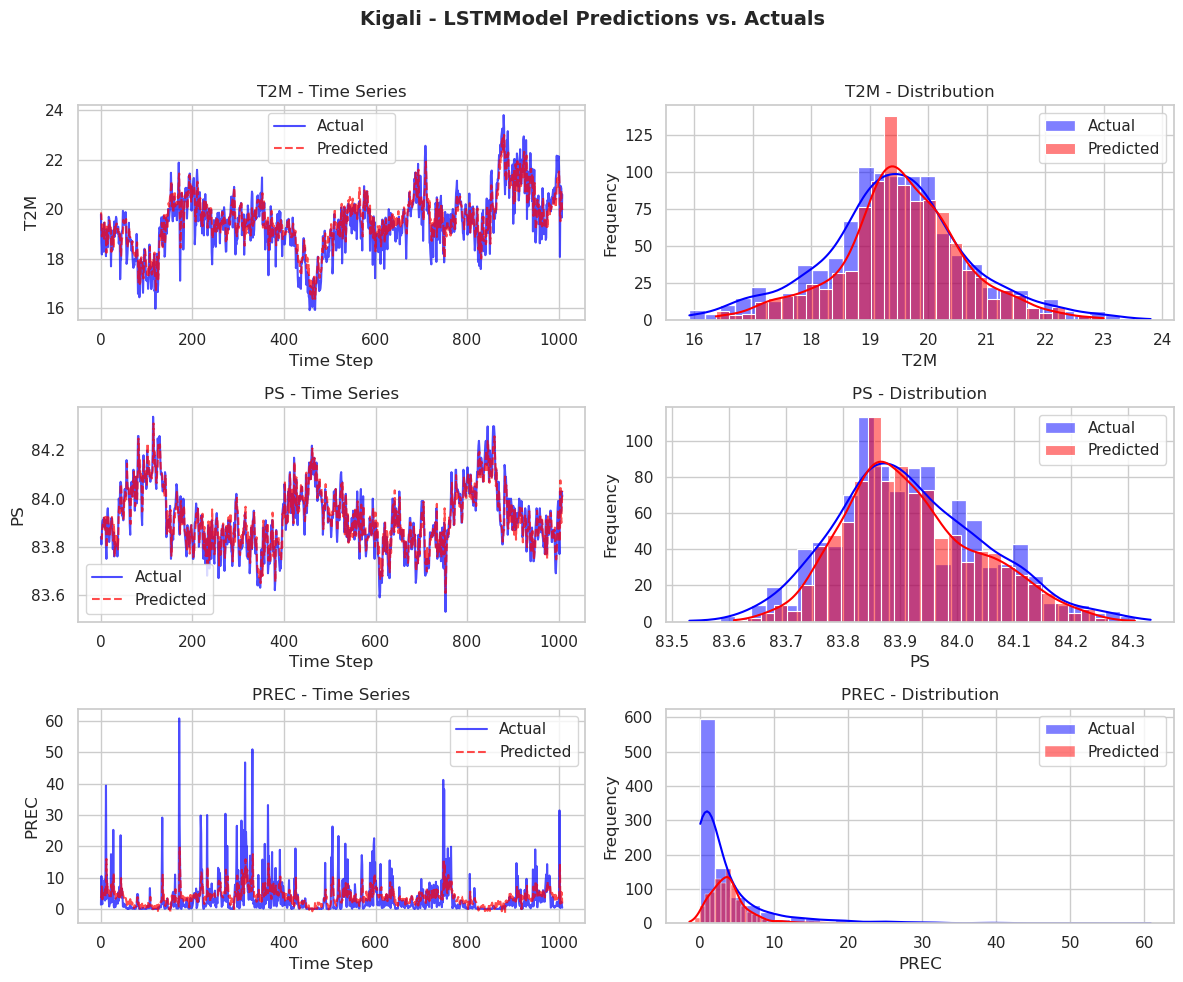}
        \caption{Abidjan and Kigali LSTM}
\end{figure}

\begin{figure}[H]
        \centering
        \includegraphics[height=6.5cm]{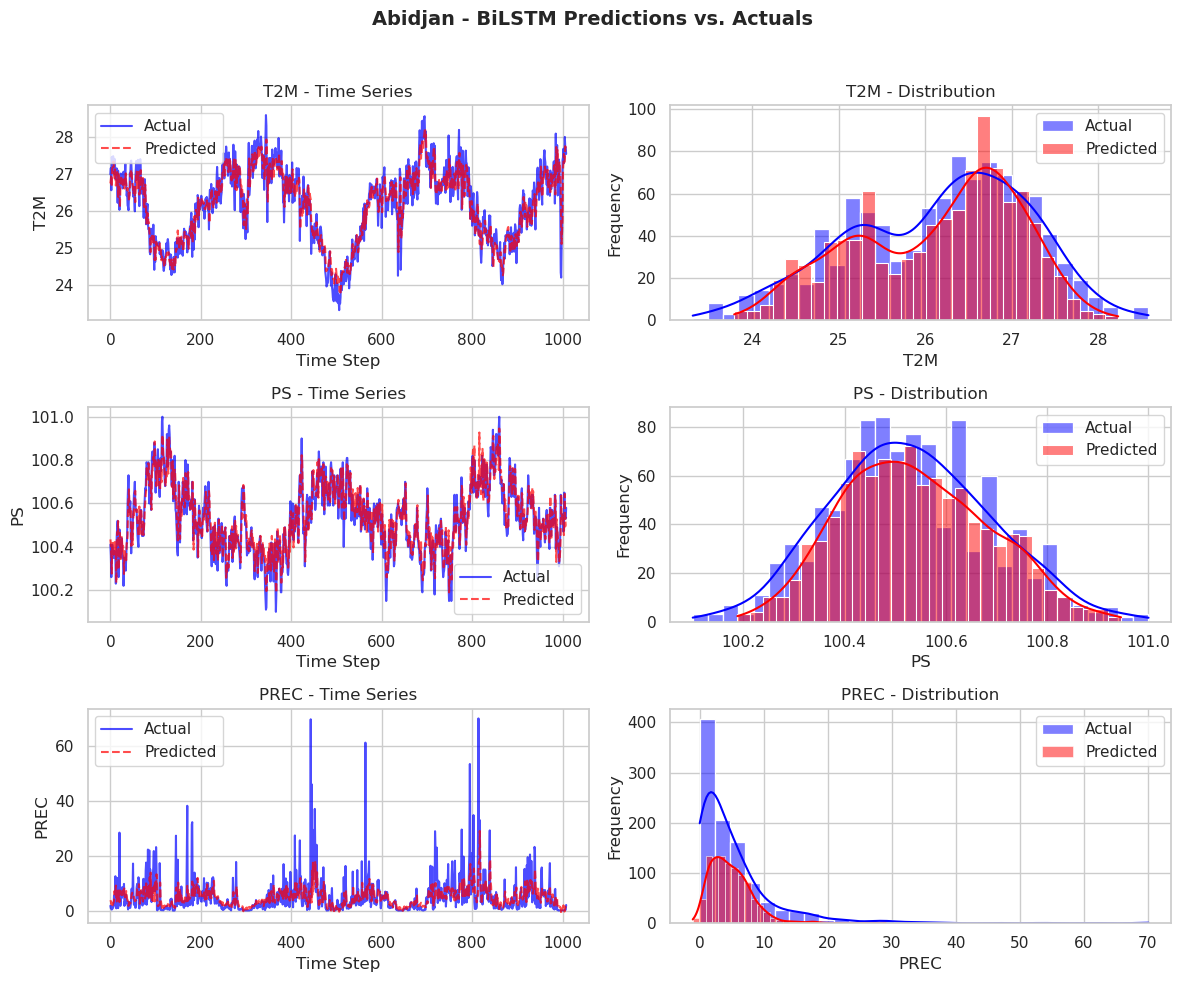}
        \includegraphics[height=6.5cm]{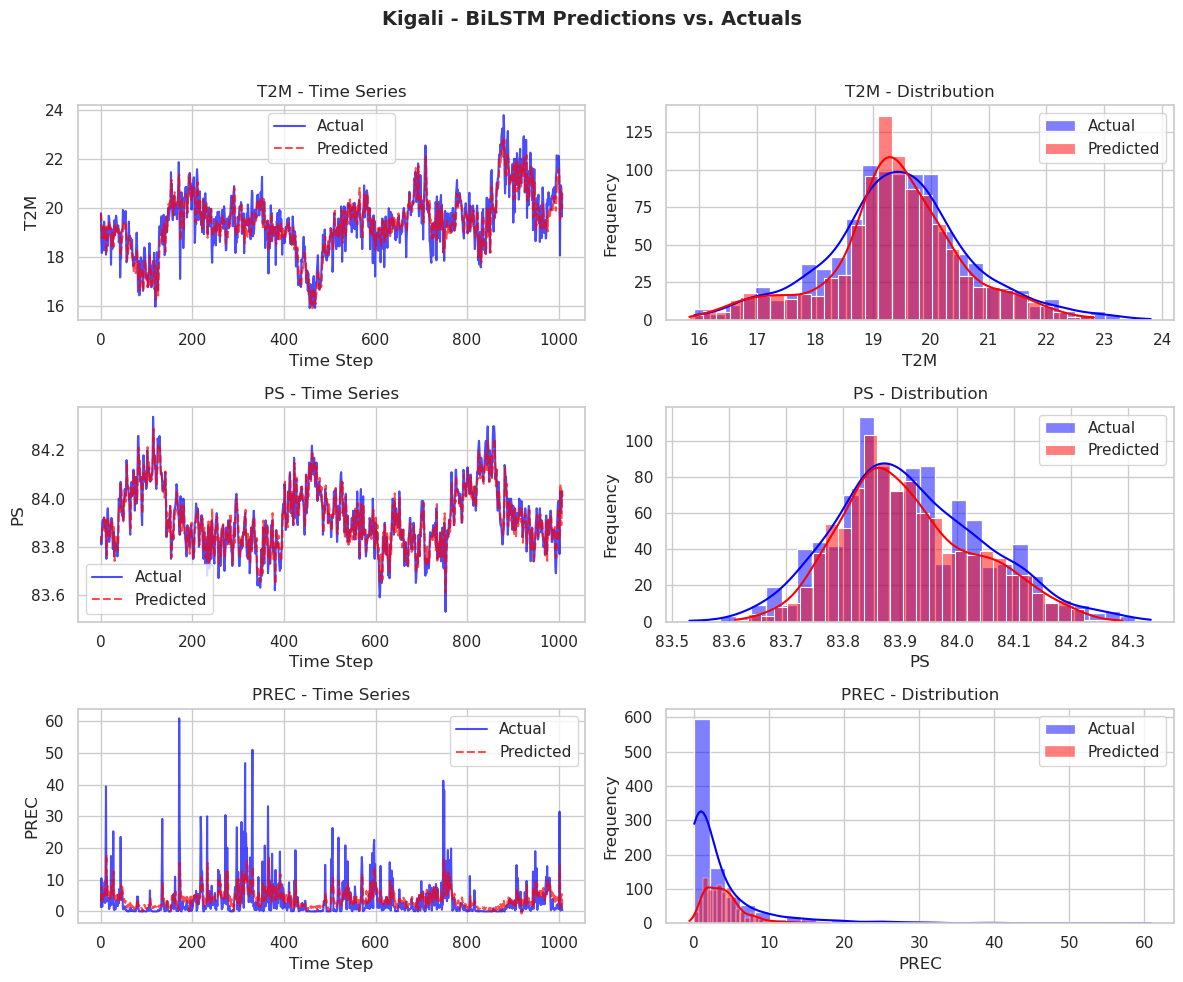}
        \caption{Abidjan and Kigali BiLTM}
\end{figure}

\begin{figure}[H]
        \centering
        \includegraphics[height=6.5cm]{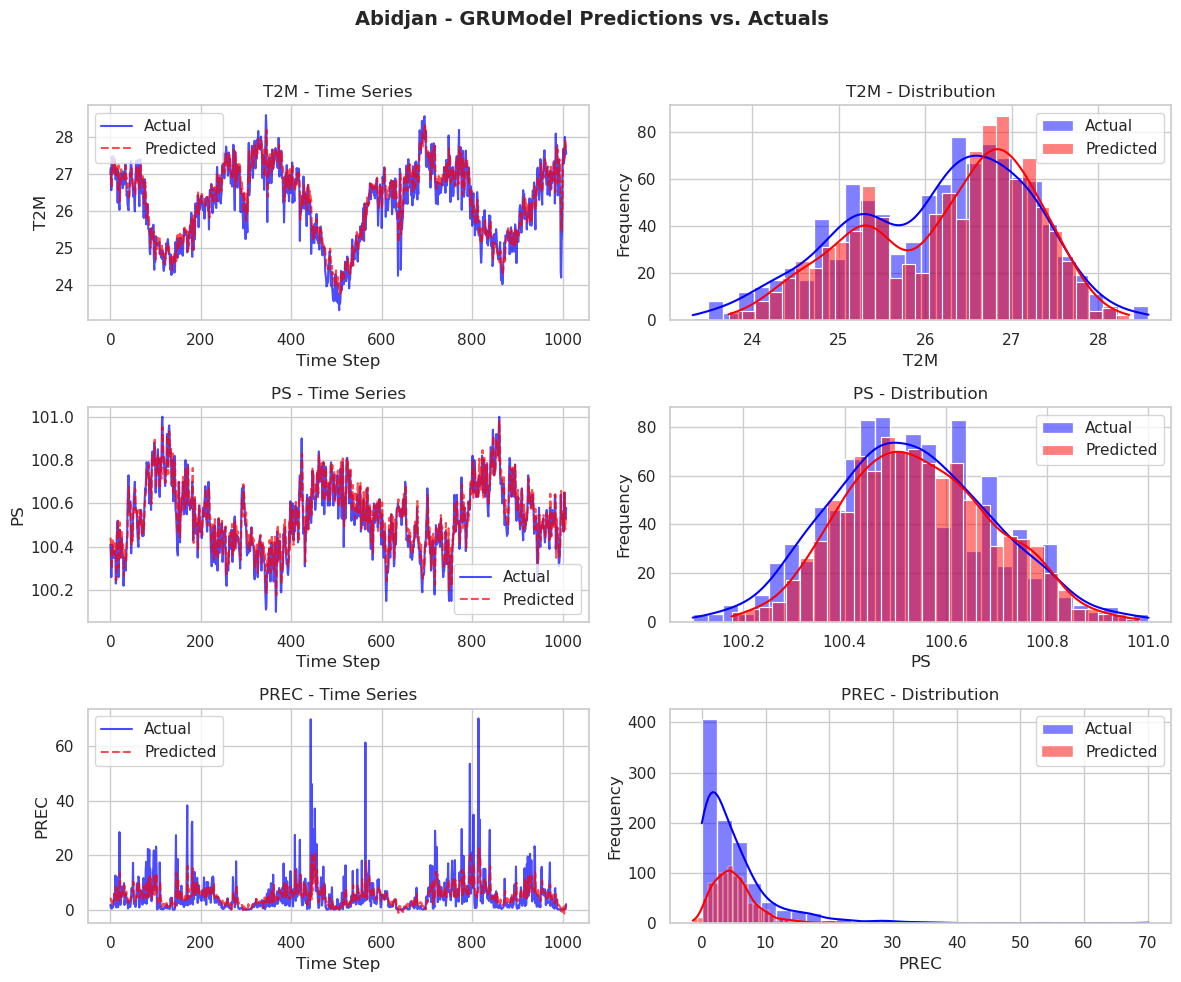}
        \includegraphics[height=6.5cm]{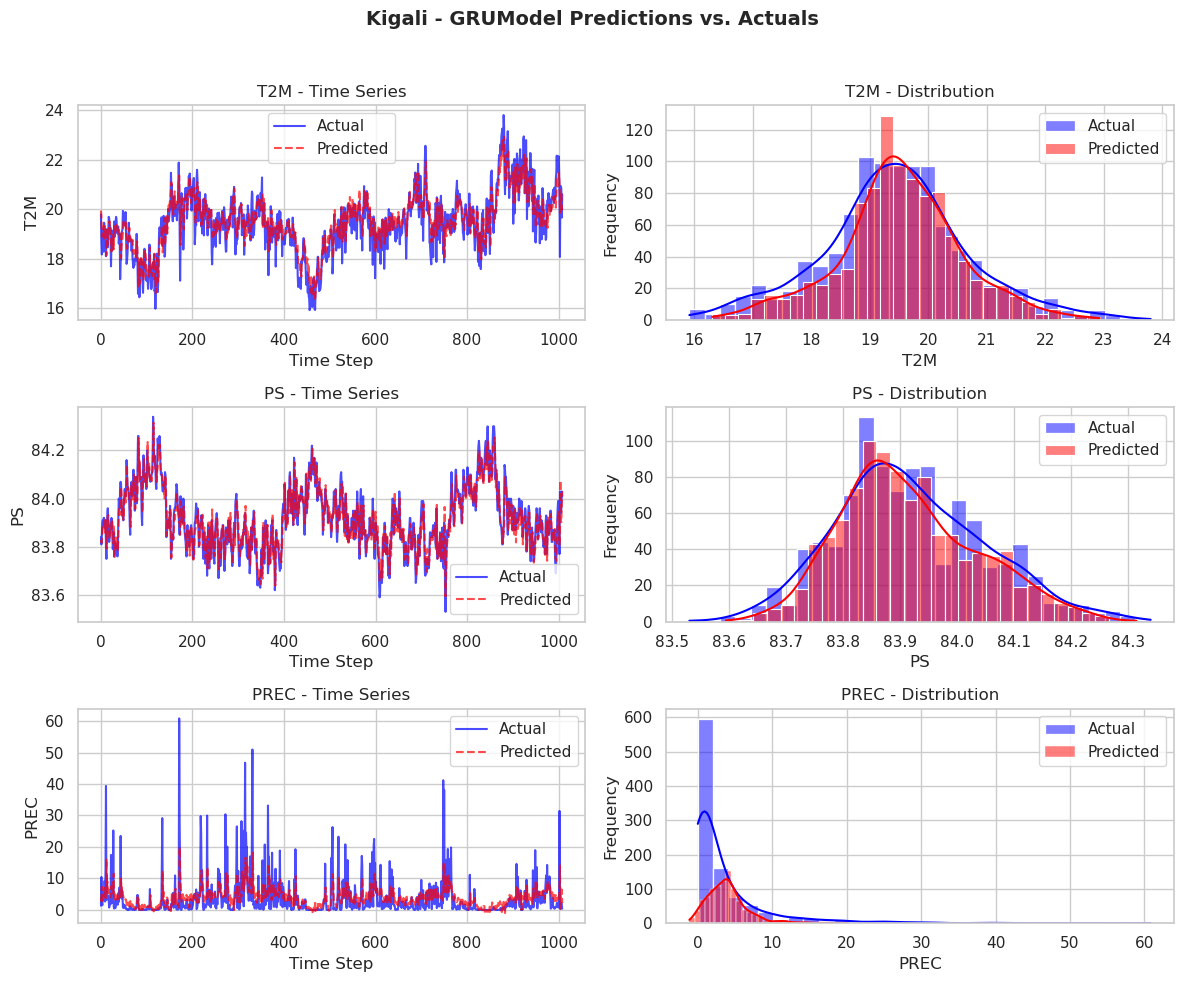}
        \caption{Abidjan and Kigali GRU}
        
\end{figure}

\begin{figure}[H]
        \centering
        \includegraphics[height=6.5cm]{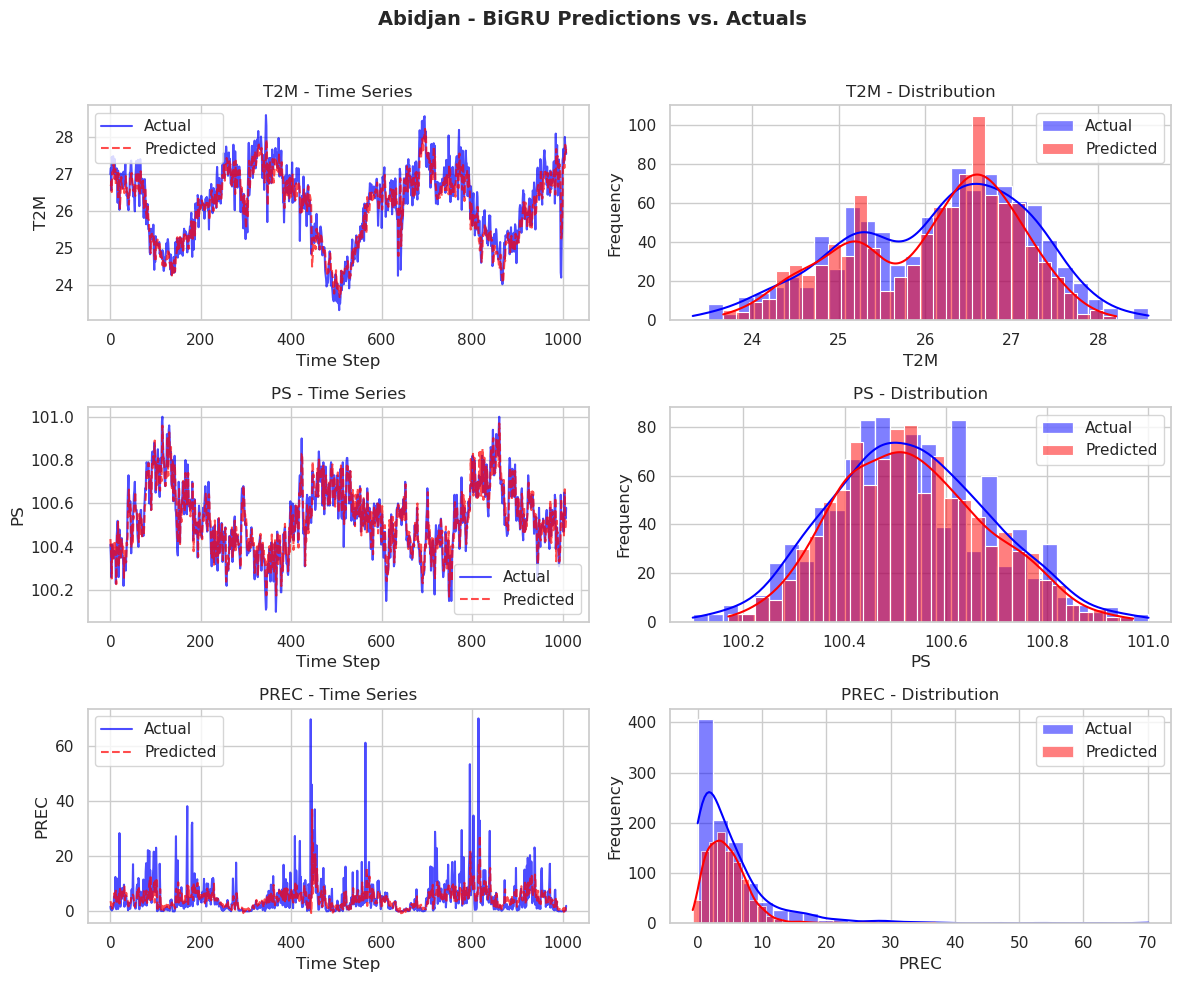}
        \includegraphics[height=6.5cm]{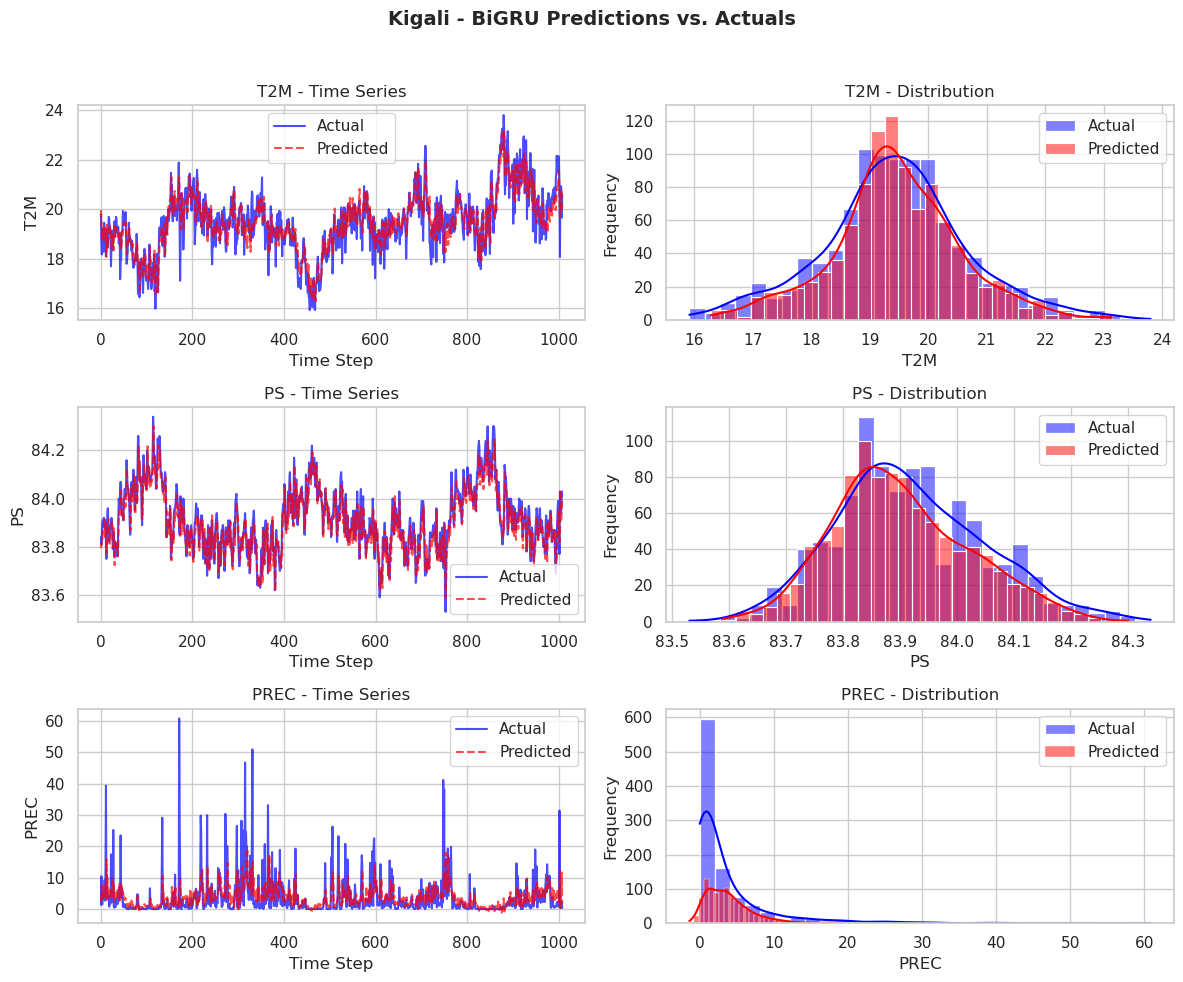}
     \caption{Abidjan and Kigali BiGru}          
\end{figure}

\begin{figure}[H]
        \centering
        \includegraphics[height=6.5cm]{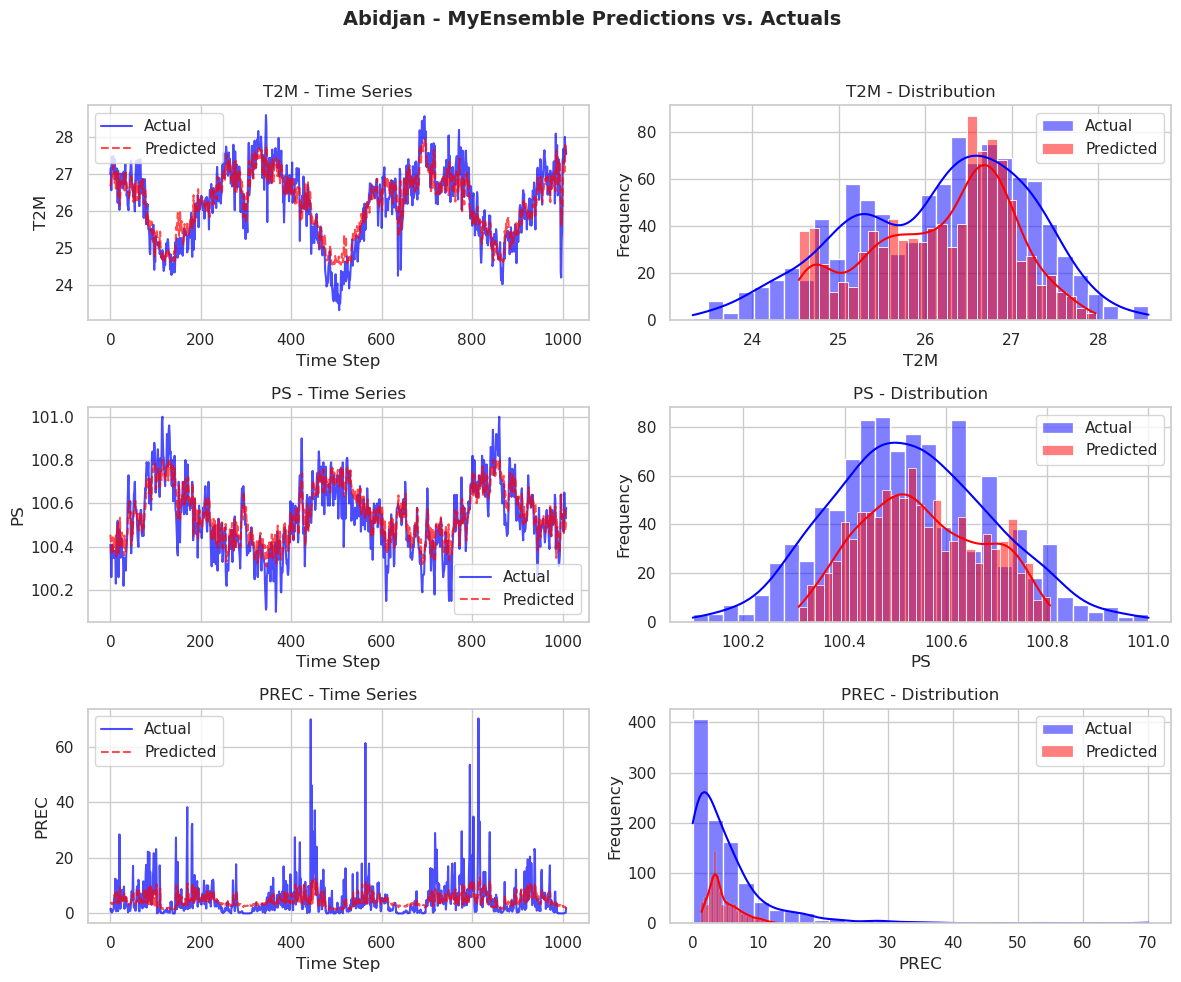}
        \includegraphics[height=6.5cm]{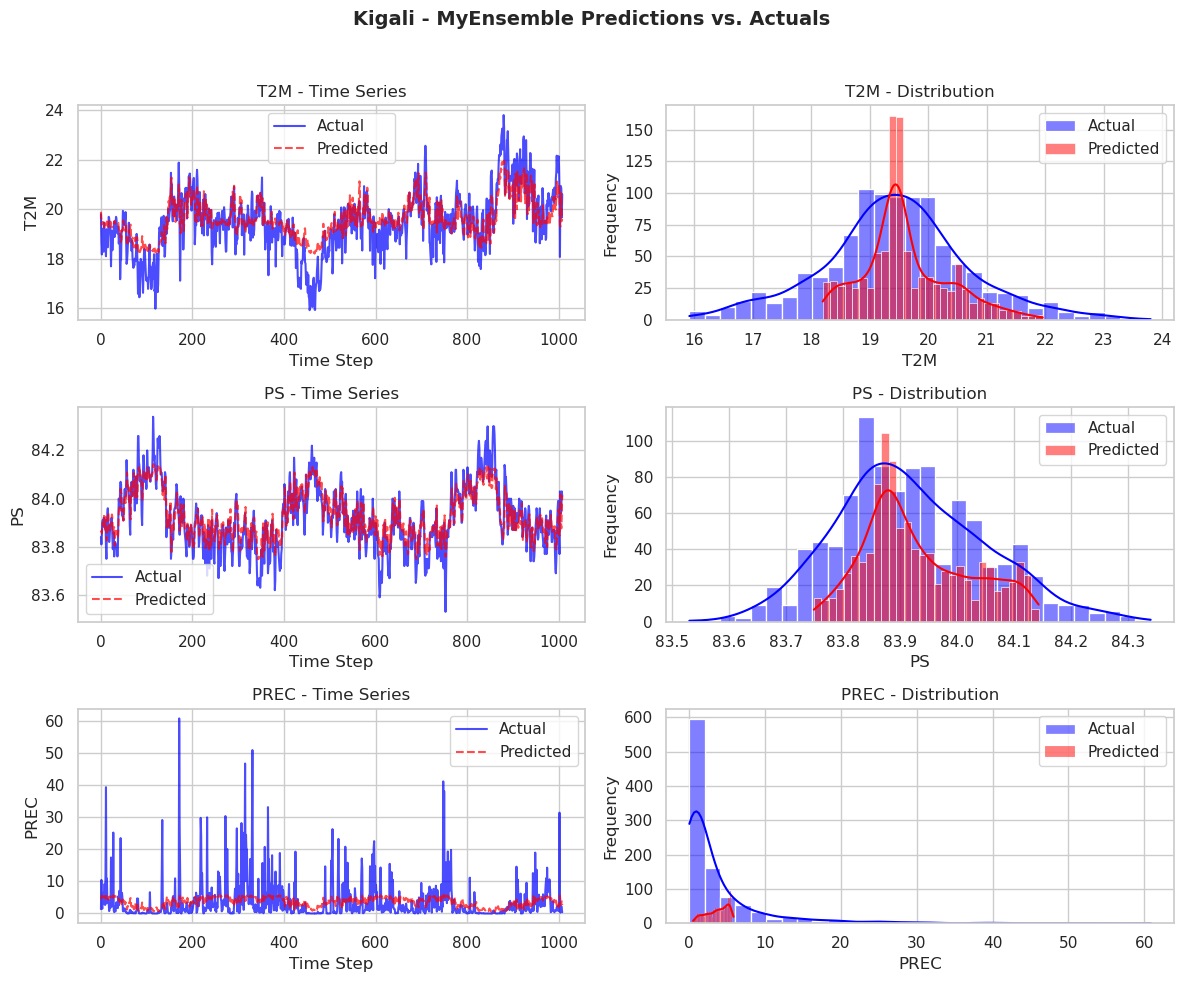}
           \caption{Abidjan and Kigali Ensemble model}     
\end{figure}
\begin{figure}[H]
        \centering
        \includegraphics[height=6.5cm]{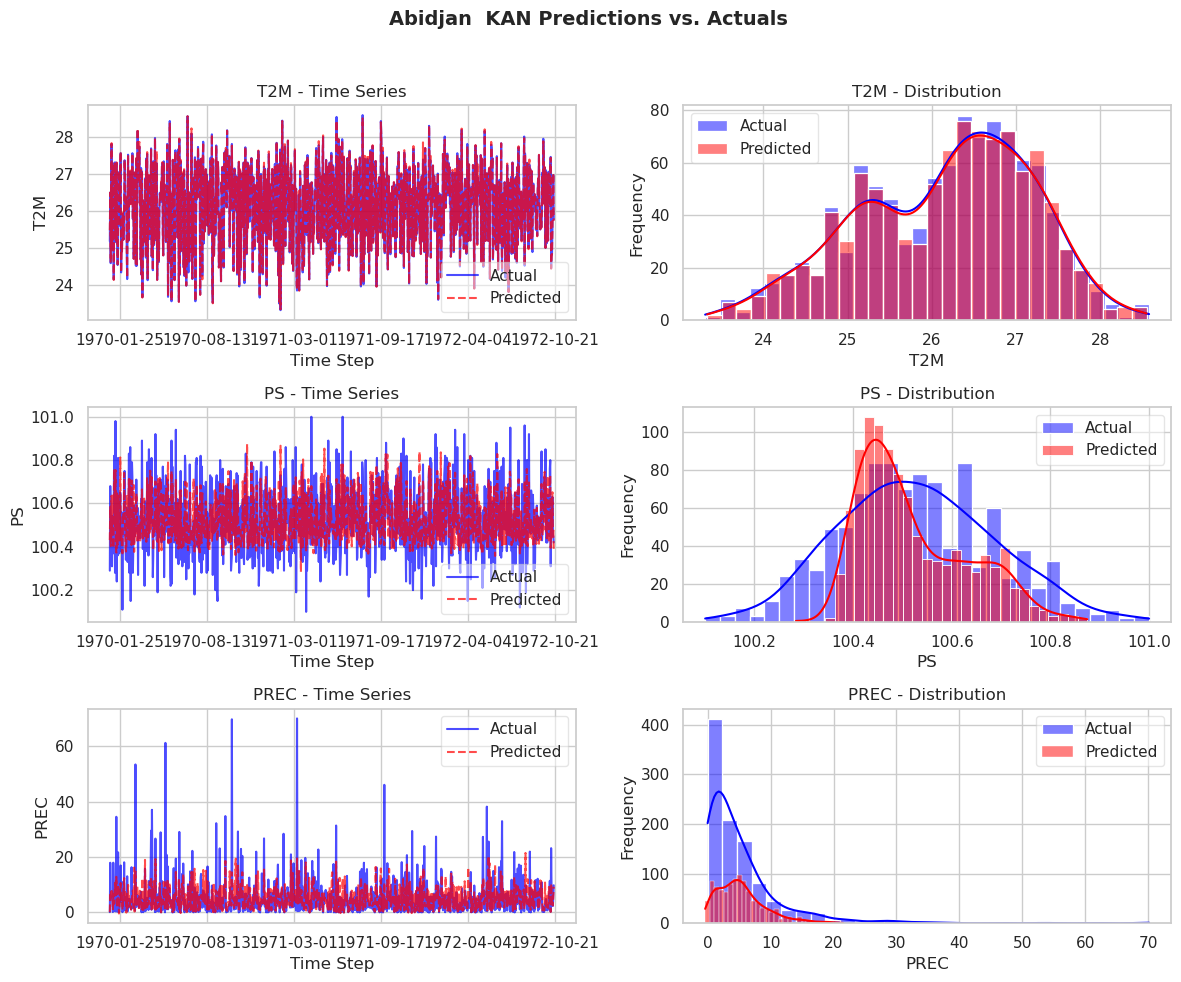}
        \includegraphics[height=6.5cm]{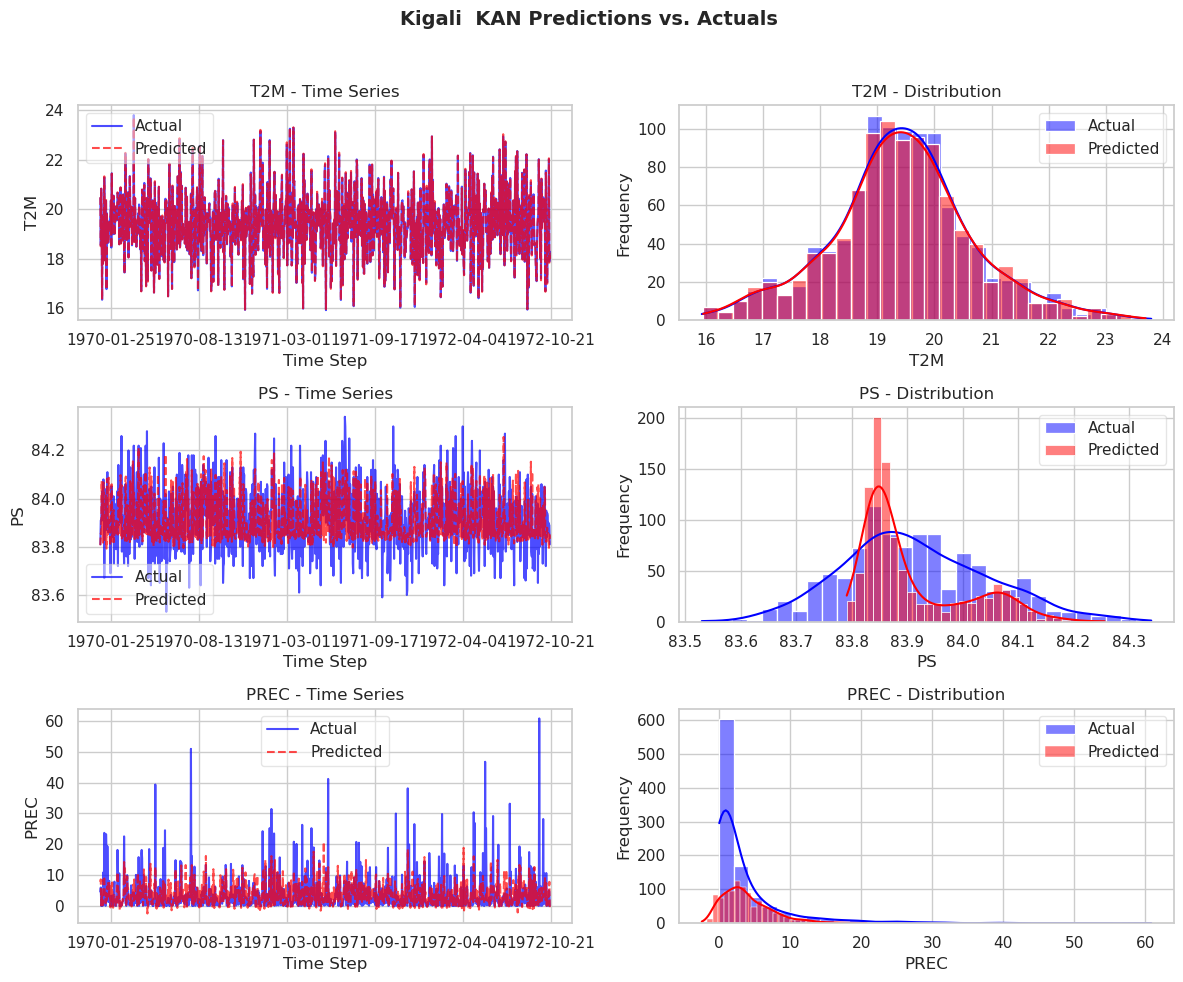}
           \caption{Abidjan and Kigali KAN}     
\end{figure}

\begin{figure}[H]
        \centering
        \includegraphics[height=3.5cm]{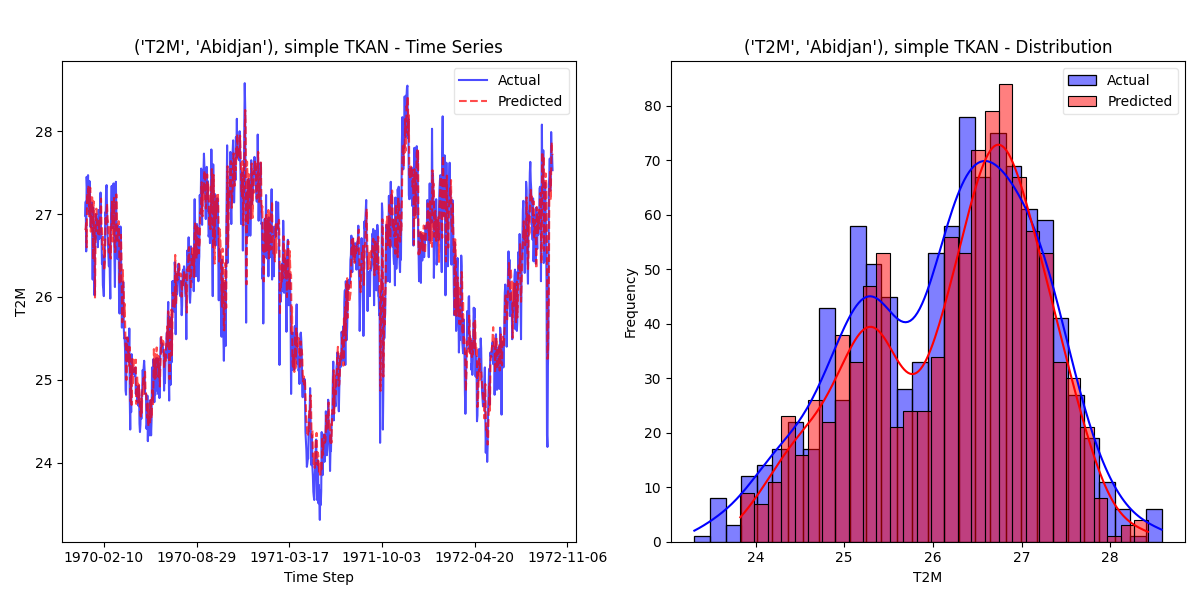}
        \includegraphics[height=3.5cm]{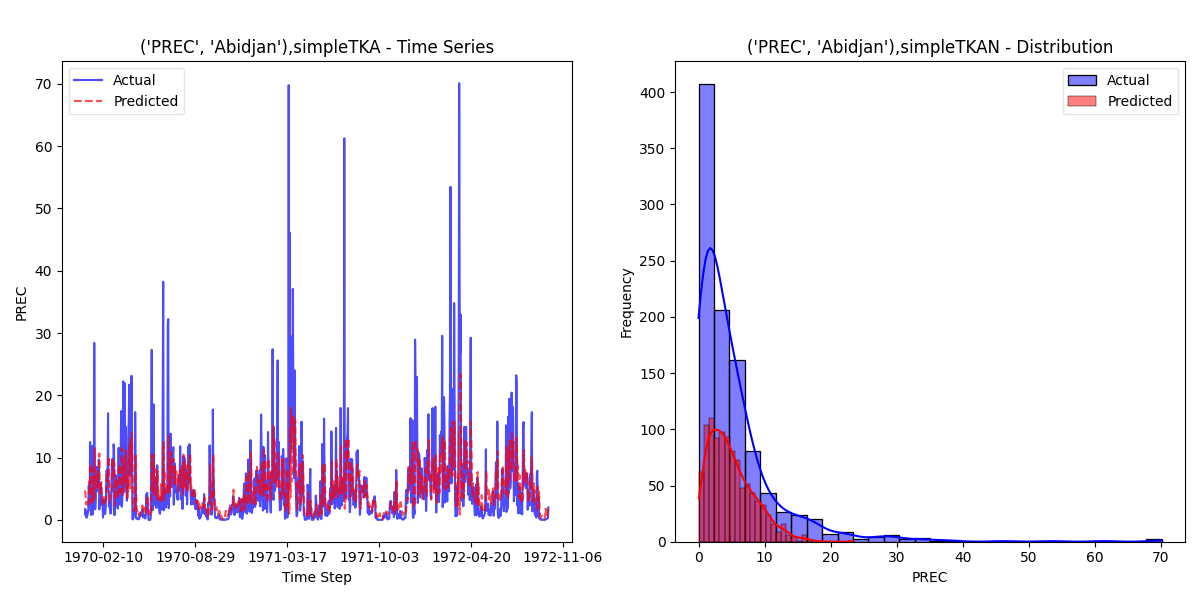}
        \includegraphics[height=3.5cm]{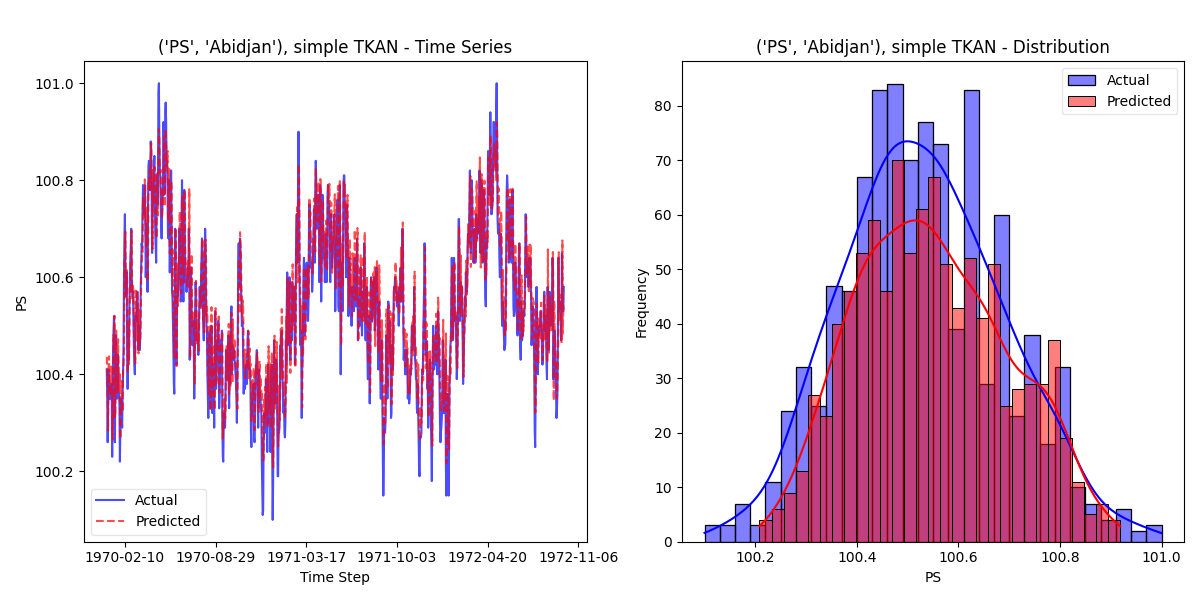} 
        \caption{Abidjan  TKAN}
        \includegraphics[height=3.5cm]{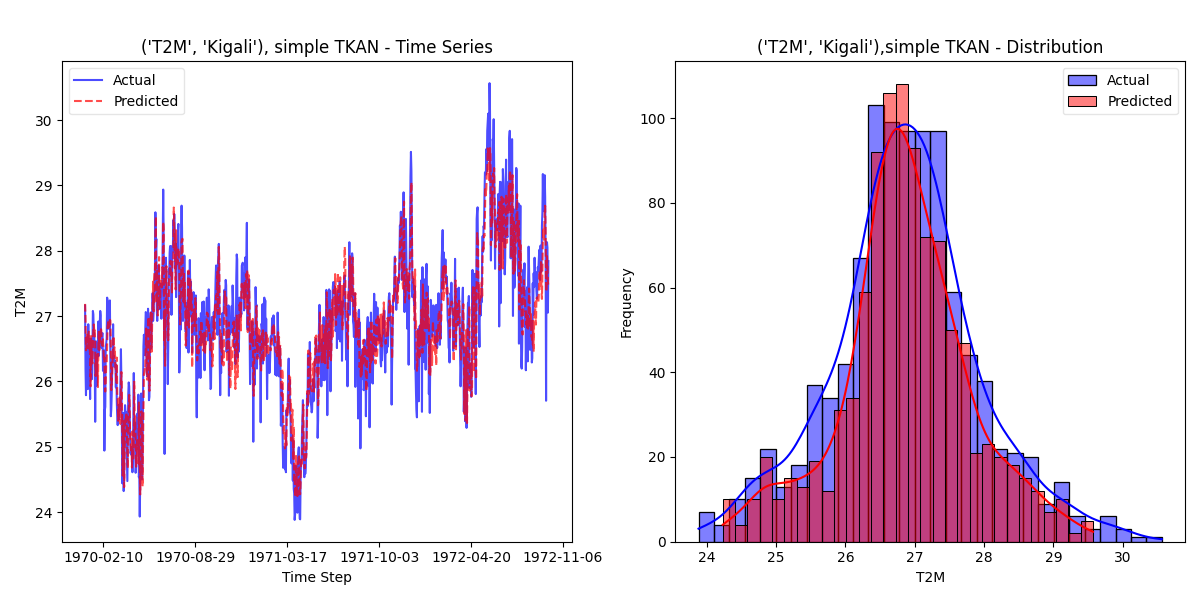} 
        \includegraphics[height=3.5cm]{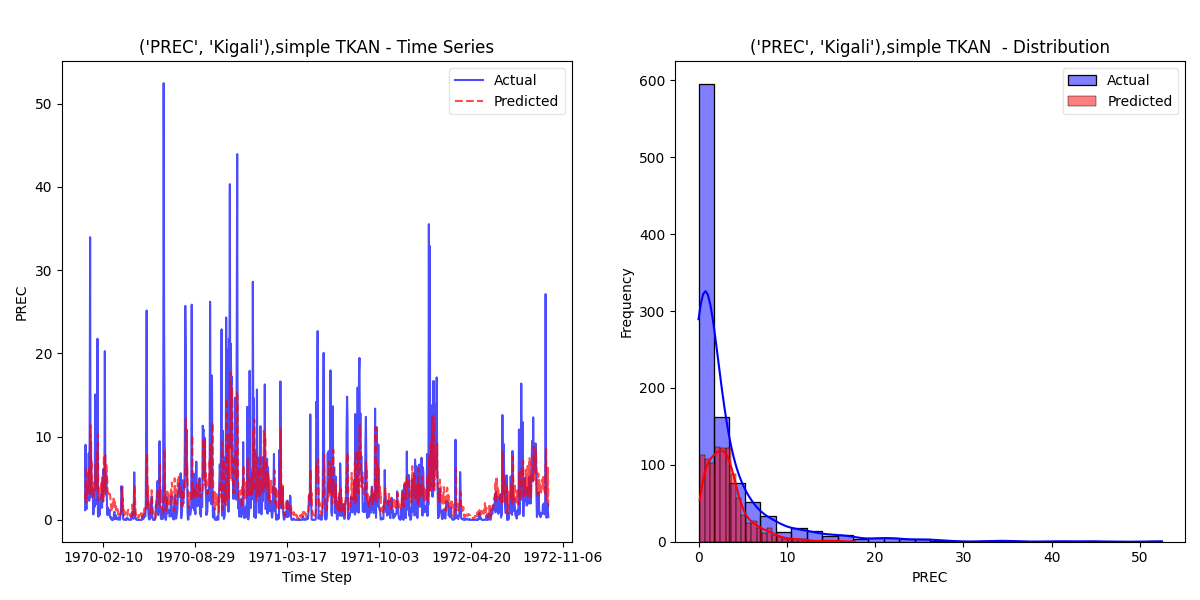} 
        \includegraphics[height=3.5cm]{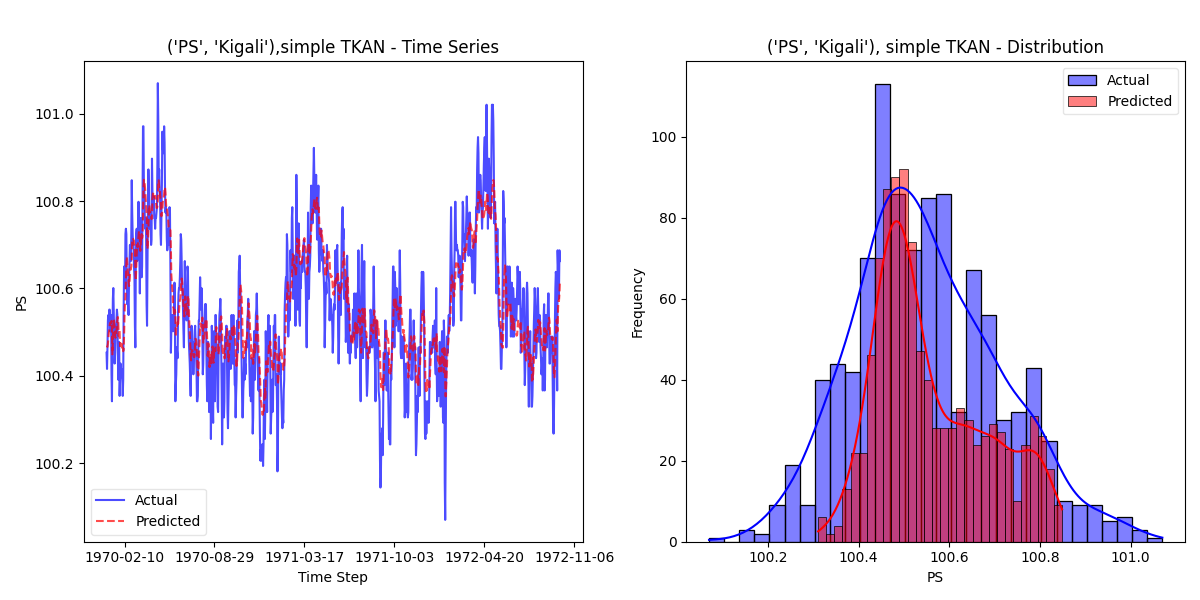} 
        \caption{ Kigali  TKAN}
\end{figure}
\clearpage

\end{document}